# A Mode-Unified, Continuous, and Biomimetic Control Strategy of a Robotic Lower Limb Prosthesis

**Hanjun Kim[1], Sixu Zhou[1,2], John Shim[1,2], Xinyuan Guo[1], Kinsey R. Herrin[1,2], and Aaron J. Young[1,2]**

**Abstract**

Existing biomimetic lower-limb prosthesis controllers are commonly designed for discrete locomotion modes, requiring separate parameterizations for level walking, ramps, and stairs. We present a mode-unified prosthesis control framework that represents knee and ankle biomechanics continuously as functions of terrain slope and gait phase across those locomotion conditions. Locally optimized ascent and descent polynomial models were unified using radial basis function (RBF) interpolation, preserving the accuracy of the partitioned models while outperforming a global polynomial representation. The resulting biomechanics were transformed into continuous and slope-dependent impedance parameters and implemented within a single finite-state-machine (FSM) controller using bilateral sensing for gait-state estimation. Experiments with 10 individuals with transfemoral amputation demonstrated terrain-adaptive locomotion from level walking through ramps and stairs up to ± 35 °, including 13 ° ramp and 15° stair conditions absent from the underlying able-bodied (AB) dataset. Experimental results reproduced the principal slope-dependent biomechanical trends predicted by the unified model, with high agreement for knee-related measures ($r^2$ = 0.82 - 0.99) and ankle kinematics ($r^2$ = 0.78 - 0.96). These findings demonstrated that unified biomechanical parameterization enables a prosthesis control architecture across diverse terrain conditions while preserving key features of AB biomechanics without requiring independently tuned impedance parameter sets for each locomotion mode. To facilitate reproducibility and further development, the biomechanics modeling and impedance parameterization code is publicly available at https://github.com/kim2576.

## 1. INTRODUCTION

The growing availability of open-source human locomotion datasets (Hood *et al.*, 2020; Camargo *et al.*, 2021; Reznick *et al.*, 2021; Scherpereel *et al.*, 2023; Wei *et al.*, 2023) spanning diverse conditions or environments has provided unprecedented insight into how lower-limb biomechanics adapt to changing terrain and task demands. These biomechanical data have been widely utilized in developing lower-limb wearable robots because reproducing human dynamics offers nominal control behaviors without costly user-specific tuning (Best *et al.*, 2023; Cortino, Best and Gregg, 2024; Lee, Lee and Young, 2024; Molinaro, Kang and Young, 2024).

Although available datasets encompass level walking, ramps, and stairs, they remain discrete and finite, so prosthesis controllers inherit that same limitation. Accordingly, extending them to new environments needs extra modeling and may produce substantial kinematic and kinetic errors in unseen conditions (Su *et al.*, 2019; Kim *et al.*, 2025). This is especially critical for prostheses because users cannot compensate with missing joints. Prior work interpolated biomechanics across adjacent ramp inclines, (Embry, Villarreal and Gregg, 2016; Embry *et al.*, 2018), stair heights (Cheng *et al.*, 2022), or walking speeds (Lenzi, Hargrove and Sensinger, 2014) improving scalability within a domain. But discontinuities remain between domains such as overground walking and stairs, which are still commonly modeled as separate tasks. Task-agnostic or unified prosthesis controllers have also been proposed to enable multiple ambulation activities without explicit terrain classification (Cowan *et al.*, 2023; Sullivan *et al.*, 2025). These approaches improve versatility across locomotion tasks but generally do not prescribe healthy individuals' joint biomechanics as continuous functions of terrain geometry. Existing methods therefore address either biomechanical fidelity within separate domains or unified control across tasks, leaving the simultaneous achievement of biomimicry and task-unified control largely unaddressed.

Gait phase estimation is another fundamental component of lower-limb prosthesis control because user-environment interaction dynamics vary continuously throughout the gait cycle. Conventional finite-state-machine (FSM) controllers partition the gait cycle into discrete states and assign corresponding control parameters to each state (Sup, Bohara and Goldfarb, 2008; Ledoux and Goldfarb, 2017; Culver *et al.*, 2018; Simon *et al.*, 2025). Among various strategies, impedance control remains widely adopted because of its simplicity, compliance, and ability to reproduce human-like

[1]Woodruff School of Mechanical Engineering, Georgia Tech, Atlanta, GA 30332-0405 USA
[2]Institute for Robotics and Intelligent Machines, Georgia Tech, Atlanta, GA 30332-0405 USA

Corresponding author:
Hanjun Kim, Woodruff School of Mechanical Engineering, Georgia Tech, Atlanta, GA 30332-0405 USA.
Email: hkim910@gatech.edu

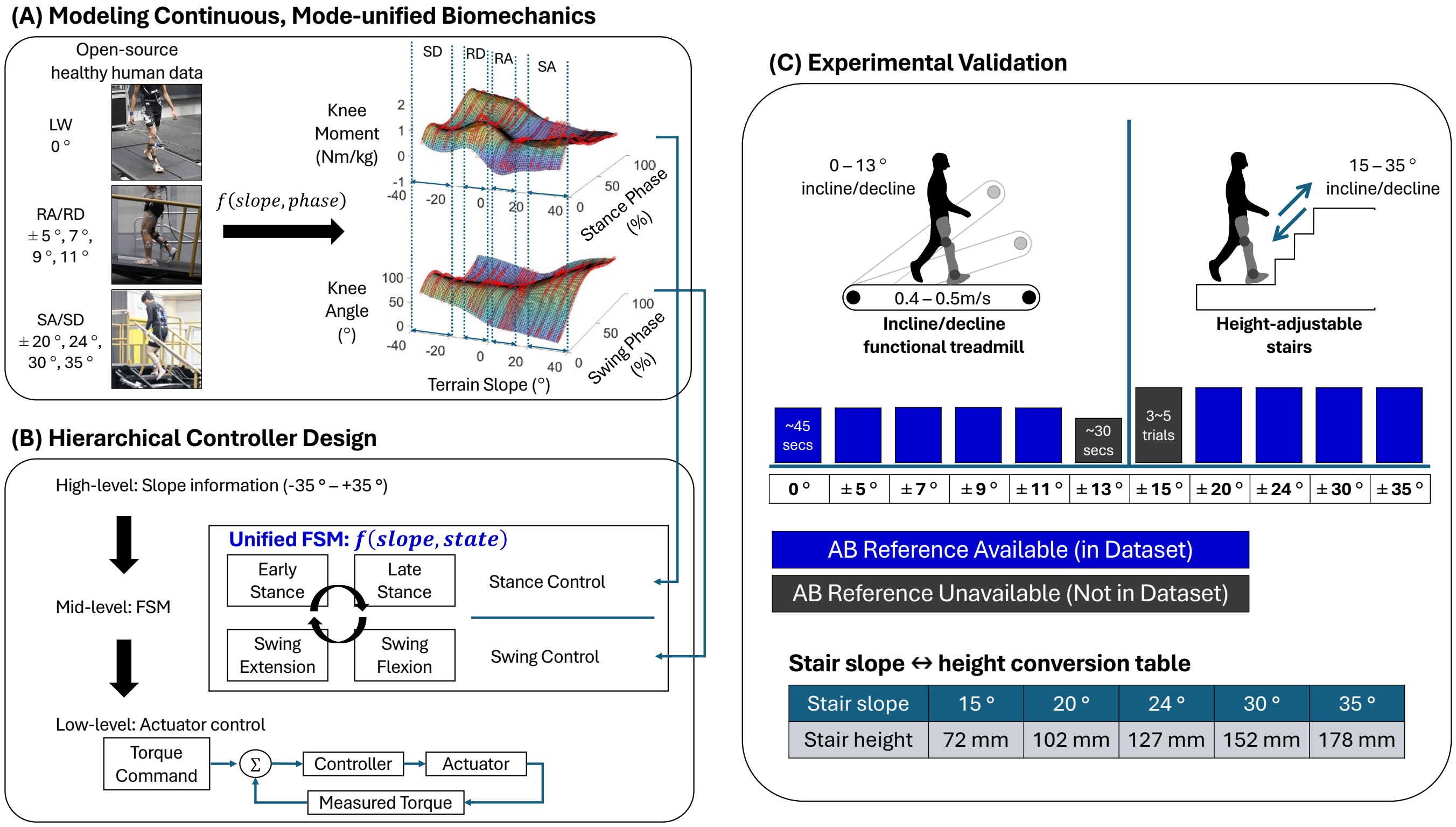

**Fig. 1.** Overview of the proposed mode-unified prosthesis control framework. **(A)** Discrete human locomotion data were transformed into continuous biomechanical models parameterized by terrain slope and gait phase. **(B)** These unified biomechanics were converted into continuous impedance parameters for prosthesis control within a unified FSM architecture. **(C)** Experimental evaluation was performed across level walking, ramps (0 ° - 13 °), and stairs (15 ° - 35 °). The available healthy human dataset did not contain the intermediate 13 ° ramp and 15 ° stair conditions, allowing assessment of the proposed framework in terrain regions not represented in reference biomechanics.

joint mechanics (Bhakta *et al.*, 2020; Camargo *et al.*, 2023). To replicate human joint behavior, impedance parameters are typically modulated according to the current gait phase (Simon *et al.*, 2014; Zhou *et al.*, 2025). However, these FSM-based control frameworks are commonly designed in a mode-specific manner, where locomotion tasks such as level walking, ramps, and stairs are separately modeled using different gait-state transition conditions. For example, transitions from early to late stance are often triggered when the ankle reaches a predefined dorsiflexion angle, requiring user-specific tuning of the transition threshold. If the angle threshold is not reached during a stride, the controller may fail to progress to the subsequent state as intended. Such threshold-dependent transitions increase tuning complexity and can limit generalizability across environments and prosthesis users. Recent studies proposed continuous gait phase estimation methods using global thigh orientation directly captured from inertial measurement units (IMUs) (Hong, Anil Kumar and Hur, 2021; Best *et al.*, 2023; Cortino, Best and Gregg, 2024; Cheng *et al.*, 2025) or model-based approaches informed by multiple sensors (Thatte, Shah and Geyer, 2019; Lee, Hong and Hur, 2021), which replace discrete state transitions with continuously evolving phase variables. Nevertheless, these approaches continue to rely on predefined ambulation-mode classifications, particularly for stair locomotion, where distinct control structures or transition conditions remain necessary. Although gait phase evolves continuously, the overall controller remains partitioned across locomotion modes.

To address these limitations, we propose a mode-unified prosthesis control framework that represents lower-limb biomechanics as continuous functions of terrain slope and gait phase. The proposed framework follows the hierarchical high-, mid-, and low-level control architecture commonly adopted in lower-limb prosthesis control (Sup, Varol and Goldfarb, 2011; Bhakta *et al.*, 2020; Camargo *et al.*, 2023). First, knee and ankle biomechanics spanning level walking, ramps, and stairs were modeled within a continuous terrain slope and gait phase domain (Fig. 1A and Section II). In particular, the stair geometry was represented as an equivalent terrain slope rather than height, to describe both ramp incline and stair within a common slope domain, as illustrated in Fig. 1C. Locally optimized models were developed separately for ascending and descending locomotion and subsequently unified while preserving their local biomechanical characteristics. Stance- and swing-phase representations were also modeled separately to retain phase-dependent locomotion dynamics. These unified biomechanical representations were then transformed into continuously slope-dependent impedance parameters for prosthesis control (Section III). Finally, the proposed framework was experimentally validated in Section IV. The resulting controller employs a bilateral sensing strategy to detect gait events from both-limb forward kinematics while maintaining a single FSM structure across all locomotion conditions. This architecture enables continuous modulation of prosthesis impedance with terrain slope without requiring mode-dependent gait-state definitions or independently tuned parameter sets. The proposed framework was experimentally evaluated across level walking, ramps, and stairs, including intermediate terrain conditions that were not explicitly represented in the underlying able-bodied (AB) biomechanical dataset (Fig. 1C). We tested two main hypotheses: **(H1)** the unified biomechanics model preserves the fidelity of the local ascent and descent fits and **(H2)** the proposed unified controller framework preserves the principal slope-dependent trends in healthy individuals' biomechanics and interpolates these trends to intermediate conditions, which are absent from the reference AB data.

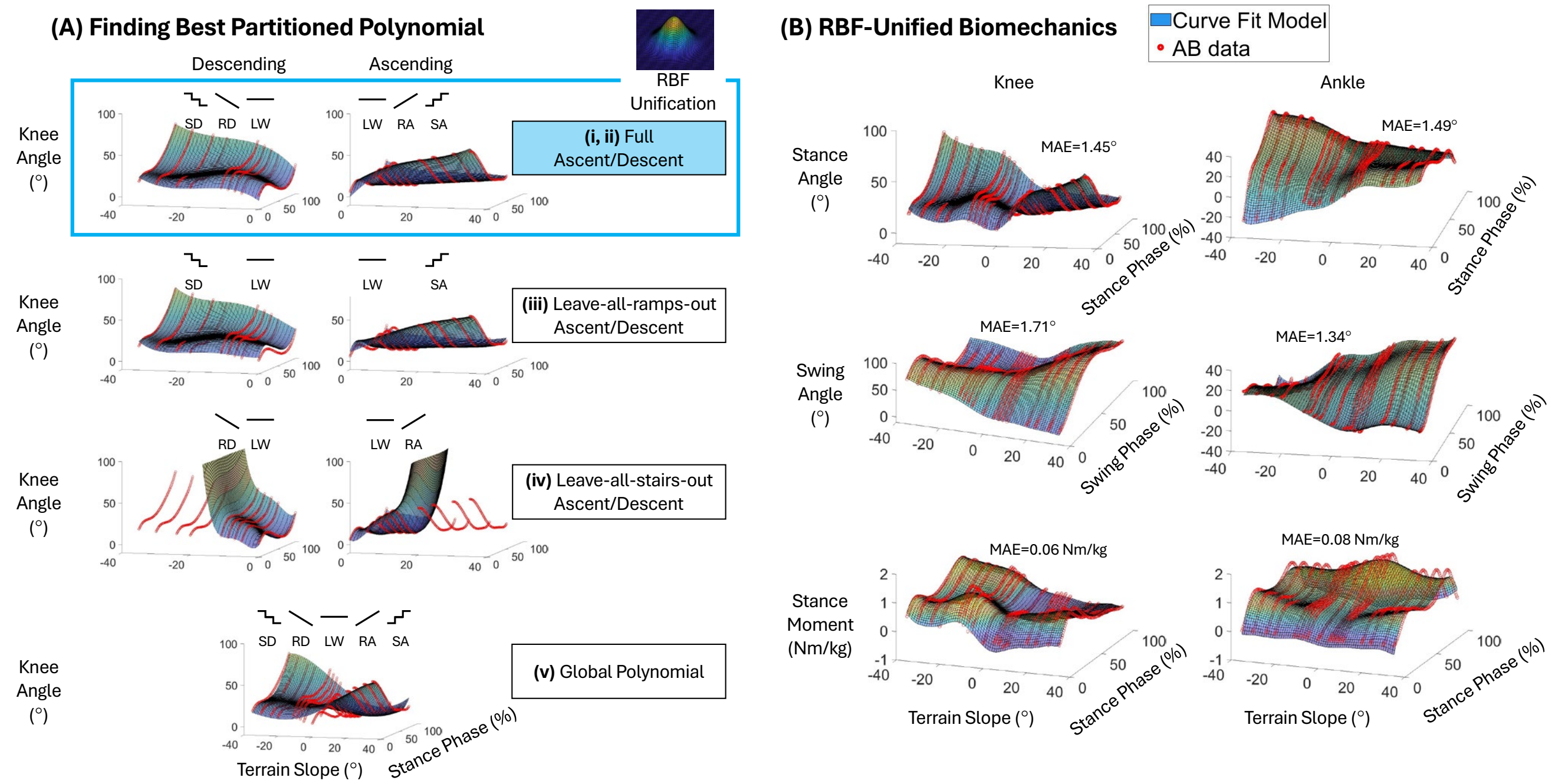

**Fig. 2.** Determining the polynomial fitting domain. **(A)** Leave-all-stairs-out fitting fails to reconstruct stair biomechanics, whereas leave-all-ramps-out fitting exhibits moderate degradation, with lower accuracy than full ascent / descent fitting (i and ii). As a result, configurations (i) and (ii) were selected as final candidates. **(B)** Representative continuous biomechanics surfaces obtained after RBF unification of the partitioned polynomial models. The red dotted lines represent the locomotion data from healthy individuals, and the surfaces represent models across various fitting configurations.

## II. Unified Locomotion Biomechanics

This section develops continuous lower-limb biomechanical models parameterized by terrain slope and gait phase. First, partitioned polynomial models are constructed separately for the ascending and descending terrain domains from open-source locomotion datasets (Camargo *et al.*, 2021; Molinaro, Kang and Young, 2024). These locally accurate representations are subsequently unified using radial basis function (RBF) interpolation, a well-established approach for constructing continuous multivariate functions from discrete observations (Micchelli, 1986; Buhmann, 2000). The resulting biomechanics surfaces serve as the foundation for the impedance parameterization framework in Section III.

### *A. Locomotion Biomechanics Dataset*

We utilized knee and ankle kinematics (joint angles) and kinetics (joint moments) collected from healthy individuals (Camargo *et al.*, 2021; Molinaro, Kang and Young, 2024). As illustrated in Fig. 1A, locomotion conditions consisted of level walking (LW, 0 °), ramp ascent/descent (RA / RD, ± 5 °, 7 °, 9 °, and 11 °), and stair ascent/descent (SA / SD, ± 20 °, 24 °, 30 °, 35 °). Four ramp and four stair conditions were selected to balance data distribution across locomotion modes. Joint angle trajectories were segmented into stance and swing phases, whereas joint moment trajectories were modeled over the stance phase only. All trajectories were phase-normalized to 101 samples. The LW trials were selected from the slow-speed condition, up to 0.6 m/s.

### *B. Determining the Polynomial Fitting Domain*

Several polynomial fitting configurations were investigated to determine an appropriate polynomial fitting domain prior to model unification (Fig. 2A). These included partitioned polynomial models for full ascent and descent, leave-ramp-out and leave-stair-out partitioning strategies, and a single global polynomial model spanning the entire terrain domain as below.

(i) Full ascent partition: LW, RA, and SA
(ii) Full descent partition: LW, RD, and SD
(iii) Leave-all-ramps-out partition: LW (0 °) and either SA or SD (± 20 °, 24 °, 30 °, 35 °)
(iv) Leave-all-stairs-out partition: LW (0 °) and either RA or RD (± 5 °, 7 °, 9 °, 11 °)
(v) Global Polynomial: LW, RA, RD, SA, and SD

Joint biomechanics $\tilde{B}_{J,D}(s,p)$ were modeled as two-dimensional fourth-order polynomial functions of terrain slope and gait phase

$$\tilde{B}_{J,D}(s,p) = \left( \sum_{l=0}^{4} \sum_{m=0}^{4-l} c_{l,m} s^l p^m \right)_{J,D} \tag{1}$$

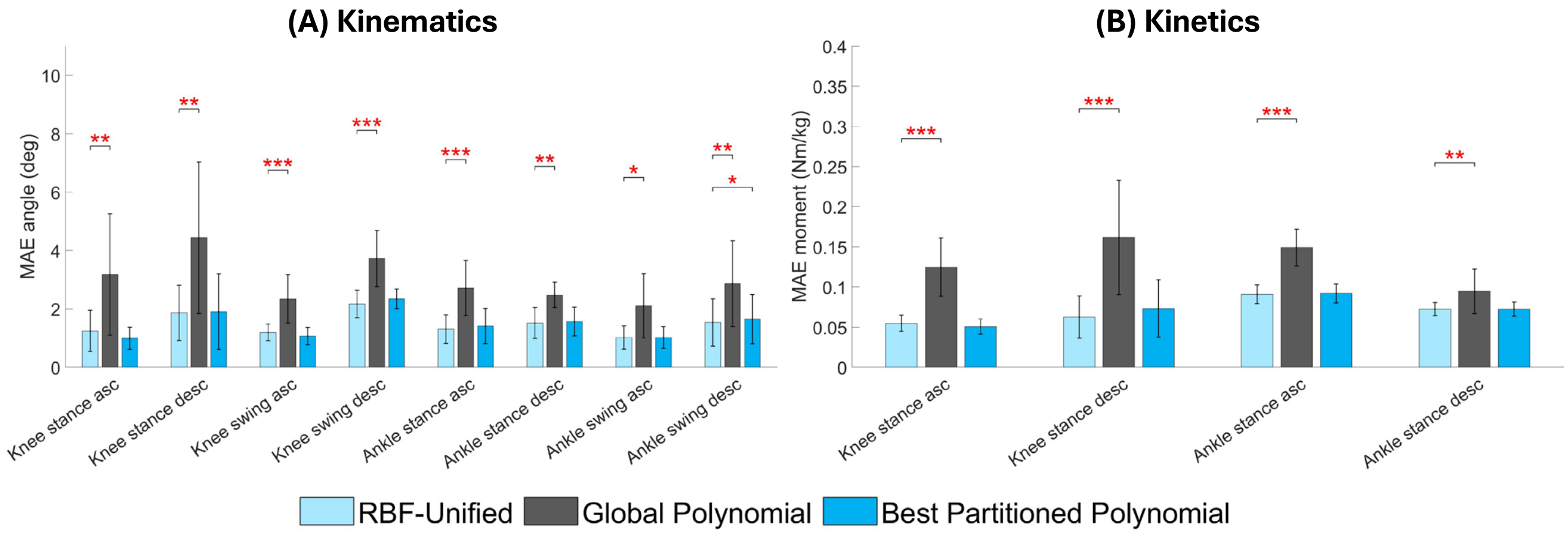

**Fig. 3.** Validation of the unified biomechanics. Proposed RBF-unified models were compared with the partitioned polynomial and global polynomials for **(A)** kinematics and **(B)** kinetics. The RBF-unified representation maintained the reconstruction accuracy of the partitioned polynomial models while consistently outperforming the global polynomial representation. Bars represent the mean ± 1 SD of the MAE across all evaluated terrain slope conditions. Statistical significance was evaluated using paired t-test (*$p < 0.05$, **$p < 0.01$, ***$p < 0.001$).

$$s \in S_s \subseteq S_U = \{-35, -34, \dots -1, 0, 1, \dots, 34, 35\} \quad (2)$$

$$p \in P = \{p_1, p_2, \dots, p_{101}\} = \{0, 0.01, 0.02, \dots, 0.99, 1\} \quad (3)$$

where $J$ and $D$ denote the target joint (knee or ankle) and the type of dynamics (angle or moment), respectively. The slope subset $S_s$ was defined according to configurations (i)-(v); for example, {0, 5, 7, 9, 11} for (iv) leave-all-stairs-out partition (ascending) or {-35, -30, -24, -20, 0} for (iii) leave-all-ramps-out partition (descending). The coefficients $c_{l,m}$ were estimated for each configuration using unconstrained linear least-squares (LSQ) regression, using the MATLAB R2024b Curve Fitting Toolbox. Based on the reconstruction performance summarized in Section II-D, the configurations (i) and (ii), the full ascent and full descent partitions, were selected as the local polynomial models for subsequent RBF unification. Configurations (iii) and (iv) were retained to evaluate cross-domain extrapolation, whereas configuration (v) served as a global polynomial benchmark.

## *C. RBF-Interpolation for Mode Unification*

The selected configurations (i) and (ii) from Section II-B were unified using Gaussian RBF interpolation (Buhmann, 2000), leveraging the $C^\infty$-continuity of the Gaussian kernel to provide a smooth transition between the two terrain partitions. This formulation reconciles trajectory mismatches at the shared level-walking condition while preserving the local biomechanical characteristics of the ascent and descent partitions.

Let $x_i = \begin{bmatrix} s_i/35 \\ p_i \end{bmatrix} \in \mathbb{R}^2$ denote a sampled terrain slope and gait phase coordinate, where terrain slope is normalized by its maximum value of 35 °. The Gaussian kernel between coordinates $x_i$ and $x_j$ can be defined as

$$\psi_{i,j} = \exp\left(-\left|x_i - x_j\right|_2^2\right) \quad (4)$$

Using this kernel, the unified biomechanics model is represented as a weighted sum of Gaussian basis functions centered at each slope-phase coordinate.

$$f(x) = \sum_{i=1}^{7272} a_i \cdot \exp(-|x - x_i|_2^2) \quad (5)$$

where $a_i$ denotes the corresponding RBF coefficient. The sampled coordinates were obtained from the full ascent and descent polynomial models, including their shared level-walking condition. Consequently, $\psi$ contains duplicated rows and columns, resulting in a singular or numerically ill-conditioned system. Hence, a diagonal regularization term with a small number $\varepsilon = 1\text{e}-8$ was introduced to ensure a stable and unique solution for the RBF coefficient vector $A = [a_1\ a_2\ \dots\ a_{7272}]^T \in \mathbb{R}^{7272\times1}$, yielding

$$A = (\psi + \varepsilon I)^{-1} M_v \quad (6)$$

where $M_v$ contains the stacked biomechanics samples from two selected partitions. The resulting RBF model was evaluated over the unified slope domain ($-35° \leq s \leq 35°$) and normalized gait phase to obtain the full biomechanics representation $f_U(s,p) \in \mathbb{R}^{71\times101}$.

The detailed formulation of sample construction, coefficient estimation, and evaluation over the unified slope domain is provided in Appendix A. The resulting continuous biomechanics surfaces are represented in Fig. 2B, and evaluated against the original AB reference in Section II-D.

## *D. Validation of the Unified Biomechanics Representation*

As illustrated in Fig. 2A, the leave-all-ramps-out models exhibited moderate degradation but retained reasonable reconstruction of the omitted ramp conditions, whereas the leave-all-stairs-out models failed to extrapolate to stair locomotion. These leave-one-domain analyses were conducted to characterize cross-domain extrapolation rather than to construct the final unified model. The full-ascent and full-descent models, configurations (i) and (ii), achieved the best reconstruction performance across the evaluated configurations and were therefore selected for RBF unification, as described in Section II-C and Appendix A.

The quantitative comparison focused on the partitioned polynomial, global polynomial, and proposed RBF-unified models using polynomials from (i) and (ii). As shown in Fig. 3, the RBF-unified models consistently achieved lower mean-absolute errors (MAEs) than the global polynomials, while preserving the accuracy of the partitioned polynomials. The only significant difference between the RBF-unified and partitioned models occurred for the ankle swing angle during descent locomotion, where the RBF-unified model achieved a lower MAE.

The resulting RBF-unified models are shown in Fig. 2B, demonstrating continuous slope-phase trajectories that closely reproduce the underlying AB biomechanics. These unified biomechanics subsequently served as the foundation for the impedance parameterization in Section III.

## III. Mode-Unified Impedance Parameters

The RBF-unified biomechanics models derived in Section II provide continuous representations of joint kinematics and kinetics over terrain slope and gait phase. This section transforms these biomechanical surfaces into a continuous impedance parameter space for mode-unified prosthesis control. Consistent with the proposed FSM framework, stance and swing phase parameters were identified separately. The following sections focus on the stance phase parameterization, which governs the primary load-bearing dynamics of locomotion.

At each terrain slope, $K$, $B$, and $\theta_{eq}$ were defined according to the impedance relationship between joint moment $\tau$, joint angle $\theta$, angular velocity $\dot{\theta}$,

$$\tau = -K(\theta - \theta_{eq}) - B\dot{\theta} \quad (7)$$

where $K$, $B$, and $\theta_{eq}$ denote stiffness, damping, and equilibrium angle, respectively. The unified biomechanics models provide the target kinematic and kinetic behaviors to be reproduced by the controller. However, the impedance parameters were not directly identified from the biomechanical models. Instead, additional design constraints were incorporated during the parameterization process to ensure compatibility with the unified FSM framework.

### A. Biomechanically Inspired Equilibrium Angle Trajectory

Previous studies identified stiffness, damping, and equilibrium angle simultaneously through optimization frameworks (Best *et al.*, 2023; Cortino, Best and Gregg, 2024). In contrast, the proposed controller prescribes biomechanically inspired equilibrium angle trajectories and subsequently identifies stiffness and damping through constrained optimization. This design follows established bio-inspired prosthesis controllers (Simon *et al.*, 2014; Culver *et al.*, 2018; Bhakta *et al.*, 2020; Camargo *et al.*, 2023) while extending their parameterization to a continuous terrain slope domain.

To construct the unified equilibrium angle trajectories, a

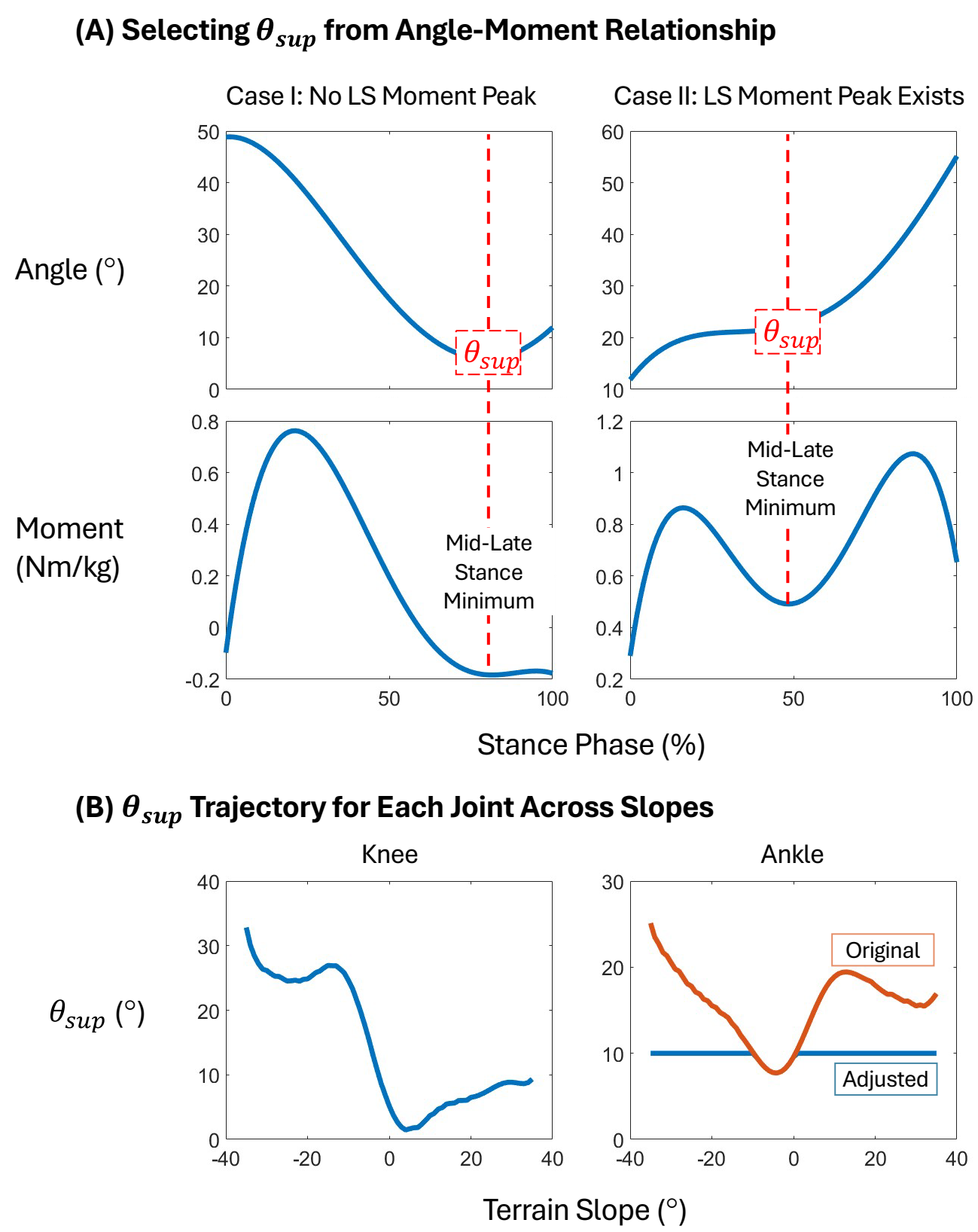

**Fig. 4.** Selection of the support equilibrium angle and resulting slope-dependent trajectories. **(A)** The support angle $\theta_{sup}$ is selected as the joint angle corresponding to the mid-to-late stance minimum of the biological joint moment. For Case I where there is no LS moment peak, this minimum occurs after the stance phase peak. On the other hand, for Case II with LS moment peak, the minimum takes place before the LS moment peak. **(B)** Continuous $\theta_{sup}$ trajectories across terrain slope. The ankle support angle is limited to 10 ° dorsiflexion to accommodate prosthesis constraints.

slope-dependent support angle ($\theta_{sup}$) was extracted from the stance phase angle-moment relationship (Fig. 4A). The $\theta_{sup}$ is defined as the joint angle corresponding to the mid-to-late stance minimum of the biological joint moment profile. Two representative cases are observed. In Case I, the biological moment does not exhibit a late stance (LS) moment peak, and the mid-to-late stance minimum occurs after the primary stance peak. This behavior is typically observed in the knee during level walking and ascending locomotion. In Case II, the biological moment exhibits a LS moment peak, and the minimum occurs before the LS peak, separating the weight-acceptance and support phase. This pattern is observed in the knee during descending locomotion and in the ankle across all evaluated environments. Based on these observations, the proposed controller uses $\theta_{sup}$ as a common biomechanical reference to initiate the support phase under both cases. In Case II specifically, rather than reproducing the initial weight-acceptance peak, the controller is designed to reproduce the support-related moment behavior following $\theta_{sup}$. This is because accurately reproducing both moment peaks would require additional modeling complexity (Schmalz, Blumentritt

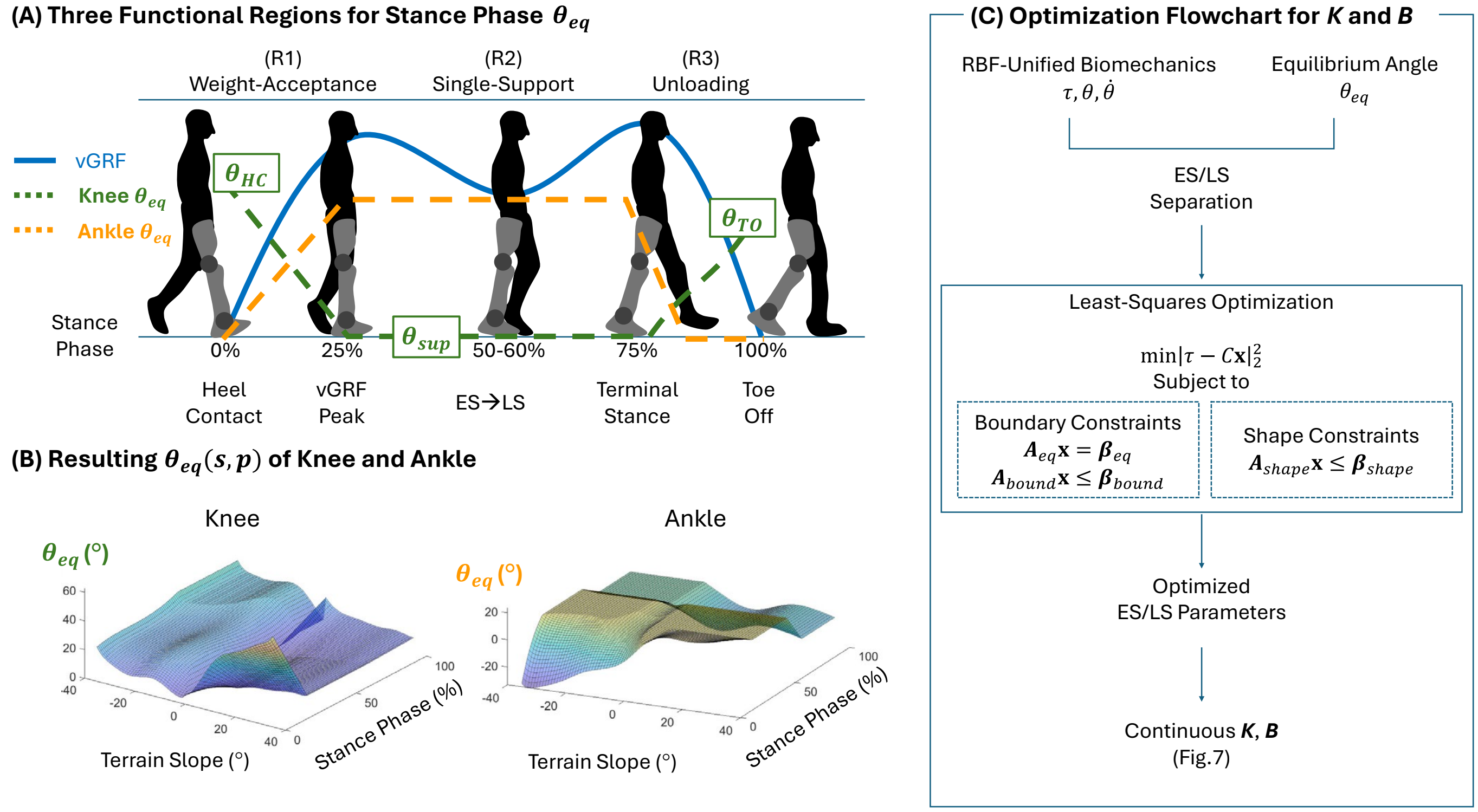

**Fig. 5.** Three-region impedance design strategy. **(A)** Schematic construction of the stance equilibrium angle trajectory. **(B)** Resulting continuous equilibrium angle surfaces across terrain slope and stance phase. **(C)** Flowchart of LSQ-optimization for *K* and *B*.

and Marx, 2007; Camargo *et al.*, 2023; Cortino, Best and Gregg, 2024).

Using $\theta_{sup}$ as the primary biomechanical reference, the stance equilibrium angle trajectory was organized into three functional regions (Fig. 5A). During weight acceptance (R1), $\theta_{eq}$ transitions from the heel-contact angle ($\theta_{HC}$) toward $\theta_{sup}$ near the 25% stance, which corresponds approximately to the first peak of the vertical ground reaction force (vGRF). During single support (R2), $\theta_{eq}$ remains at $\theta_{sup}$ so that increasing joint flexion produces support through impedance. Finally, during unloading (R3), $\theta_{eq}$ transitions from $\theta_{sup}$ toward the terminal toe-off angle ($\theta_{TO}$) to facilitate ankle push-off and knee swing initiation.

Previous powered knee-ankle prosthesis controllers have implemented these functions through load-reactive $\theta_{eq}$ modulation, with the ankle $\theta_{eq}$ reaching its terminal plantarflexion before the knee completes its transition toward flexion (Fey *et al.*, 2014; Ingraham *et al.*, 2016). This joint-specific design reflects the distinct functions of the knee and ankle during late stance. Ankle plantarflexion support must be delivered while the foot remains loaded, whereas knee flexion prepares the limb for swing phase. Accordingly, the ankle $\theta_{eq}$ was designed to reach its terminal plantarflexion before toe-off, while the knee $\theta_{eq}$ gradually approaches its flexed position near toe-off to maintain the late-stance support. The onset and duration of these transitions were independently varied during offline impedance optimization, with the selected configurations detailed in Appendix B.

For the knee, level walking and ascending locomotion directly follows the three-region structure described above. During descending locomotion, however, the transition in R1 was realized using a P-value strategy similar to previous prosthesis controllers (Simon *et al.*, 2014),

$$\theta_{eq} = \theta P(s)$$
$$P(s) = \begin{cases} 0.70 - \dfrac{0.29}{35}s, & s \leq 0 \\ 0, & s > 0 \end{cases} \tag{8}$$

As the descent becomes steeper, increasing $P(s)$ causes $\theta_{eq}$ to follow the measured knee angle more closely during R1, permitting greater controlled knee flexion before converging to $\theta_{sup}$. Once $\theta_{eq}$ reaches $\theta_{sup}$, it remains fixed throughout R2 so that additional knee flexion generates higher torque under the impedance relationship. During R3, $\theta_{eq}$ transitions toward $\theta_{TO}$ in preparation for swing.

The ankle follows the same three-region framework but incorporates two prosthesis-specific modifications. First, $\theta_{sup}$ was bounded to a maximum dorsiflexion angle of 10 ° (Fig. 4B) to accommodate the prosthetic ankle range of motion (± 30 °) and preserve sufficient plantarflexion torque. This limit also accounts for the reduced forward progression commonly observed by prosthesis users during stair ascent (Aldridge, Sturdy and Wilken, 2012; Zhou *et al.*, 2025). Consequently, the bounded $\theta_{sup}$ preserves the slope-dependent trend while remaining suitable for prosthesis control. During descending locomotion, the transition in R1 is similarly implemented using the P-value strategy,

$$\theta_{eq} = (\theta - \theta_{sup})P + \theta_{sup} \tag{9}$$

$P = 1.1$ was selected, causing $\theta_{eq}$ to follow the measured ankle angle while converging toward $\theta_{sup}$, thereby providing controlled plantarflexion yielding during weight acceptance. Once $\theta_{eq}$ reaches $\theta_{sup}$, it remains fixed throughout R2, allowing the ankle to provide stable support during single support. During R3, $\theta_{eq}$ transitions from $\theta_{sup}$ toward $\theta_{TO}$ to facilitate prosthetic push-off. Applying the above design principles across the continuous terrain slope domain yields the unified equilibrium angle trajectories for the stance phase ($\theta_{eq}(s,p)$) shown in Fig. 5B.

The experimental realization of the proposed design of equilibrium angle $\theta_{eq}$ using the support angle $\theta_{sup}$ is further illustrated in Fig. 6. For Case I (Fig. 6A), $\theta_{eq}$ approaches $\theta_{sup}$ while the knee remains flexed, generating an angular displacement that contributes to the stance knee extension. For Case II (Fig. 6B), the proportional yielding strategy allows controlled knee stance flexion before approaching $\theta_{sup}$, after which further flexion generates a resisting extension torque. These examples demonstrate how the proposed $\theta_{eq}$ trajectories contribute to desired joint behaviors, providing the foundation for the stance-phase impedance parameterization.

For the swing phase, equilibrium angles were defined from two reference points of the RBF-unified kinematic models. The swing flexion state used the maximum flexion angle to promote sufficient limb shortening and ground clearance, whereas the swing extension state used the heel-contact angle ($\theta_{HC}$) to prepare the limb for stable ground contact. Extending these references across the continuous terrain slope domain yielded unified swing phase equilibrium trajectories consistent with the stance phase design philosophy.

## B. Stance Phase Parameterization

The stance-phase stiffness and damping trajectories were parameterized using the same three-region structure introduced for the equilibrium angle (Fig. 5A). Both parameters increase during weight acceptance (R1), remain approximately constant throughout single support (R2), and remain constant or decrease during unloading (R3). This structure translates the biomechanical design principles into continuous impedance trajectories compatible with the proposed unified FSM framework.

To obtain compact impedance representations suitable for optimization and real-time implementation, both stiffness ($K$) and damping ($B$) were represented as fourth-order polynomials of normalized stance phase for each terrain slope ($s$). The overall optimization procedure is illustrated in Fig. 5C. For each joint, the early-stance (ES) impedance parameters were optimized first using the RBF-unified joint kinematics, kinetics, and predefined $\theta_{eq}$ trajectories defined in Section III-A. The resulting terminal ES parameters were then used to initialize the late-stance (LS) optimization, maintaining continuity between the two intervals. The ES-to-LS boundaries were aligned with the joint-specific onset of R3 described in Section III-A, with selected configurations detailed in Appendix B. The polynomial

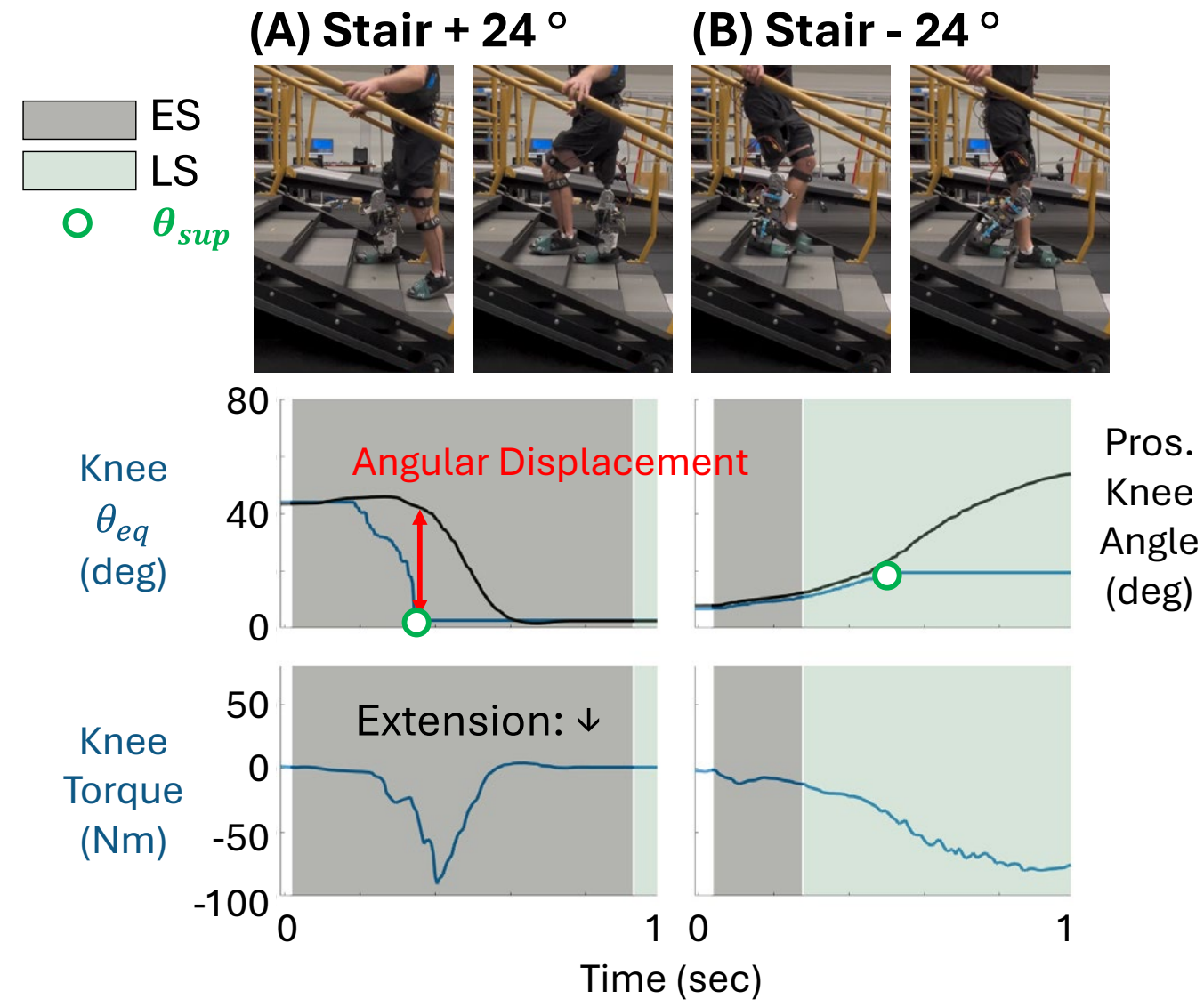

**Fig. 6.** Experimental realization of the support angle ($\theta_{sup}$, green circle) of prosthetic knee during stair ascent and descent. The upper panels compare prescribed equilibrium angle ($\theta_{eq}$, blue curves) and measured prosthetic knee angle (black curves) to highlight the angular displacement (red arrow) related to impedance rule. The lower panels illustrate the corresponding knee torques during **(A)** 24 ° stair ascent (CASE I) and **(B)** 24 ° stair descent (CASE II). During stair ascent, $\theta_{eq}$ approaches $\theta_{sup}$ while the knee remains flexed, establishing an angular displacement that generates stance knee extension. During stair descent, $\theta_{eq}$ initially follows the knee angle during weight acceptance before converging to $\theta_{sup}$. Subsequent knee flexion after the convergence generates a resisting extension torque, facilitating controlled knee yielding under load.

representation of stiffness and damping are expressed as follows.

$$K(s,p) = \sum_{i=0}^{4} k_{s_i} p^i, \qquad B(s,p) = \sum_{i=0}^{4} b_{s_i} p^i \tag{10}$$

Substituting Eq. (10) into the impedance relationship Eq. (7) yields a linear least-squares problem,

$$\begin{gathered} \min_{\mathbf{x}} |\boldsymbol{\tau}^{\boldsymbol{q}} - \boldsymbol{C}^{\boldsymbol{q}}\mathbf{x}^{\boldsymbol{q}}|_2^2 \\ \text{subject to} \\ \boldsymbol{A}_{\mathrm{eq}}^{\boldsymbol{q}}\mathbf{x}^{\boldsymbol{q}} = \boldsymbol{\beta}_{\mathrm{eq}}^{\boldsymbol{q}}, \\ \boldsymbol{A}_{\mathrm{ineq}}^{\boldsymbol{q}}\mathbf{x}^{\boldsymbol{q}} \le \boldsymbol{\beta}_{\mathrm{ineq}}^{\boldsymbol{q}} \end{gathered} \tag{11}$$

where $\boldsymbol{q} \in \{ES, LS\}$ denotes the early- or late-stance optimization interval. The coefficient vector $\mathbf{x}^{\boldsymbol{q}}$ contains the polynomial coefficients of stiffness and damping, while $\boldsymbol{C}^{\boldsymbol{q}}$ is constructed from the RBF-unified joint kinematics and prescribed equilibrium angle trajectories. The target moment vector $\boldsymbol{\tau}^{\boldsymbol{q}}$ is obtained from the RBF-unified joint kinetics. Equality and inequality constraints regulate the magnitude and progression shape of impedance parameters throughout the stance phase.

Increasing terrain steepness requires greater joint support during ascent and greater energy dissipation during descent. Therefore, slope-dependent upper envelopes, $K_{\max}(s)$ and $B_{\max}(s)$, were defined using arctangent functions connecting experimentally validated impedance parameters from 35° stair-ascent and stair-descent controllers (Bhakta *et al.*, 2020; Camargo *et al.*, 2023). Constant lower bounds, $K_{\min}$ and $B_{\min}$, were selected as empirical safety limits, yielding

$$\begin{aligned} K_{\min} \leq K(s,p) \leq K_{\max}(s) \\ B_{\min} \leq B(s,p) \leq B_{\max}(s) \end{aligned} \tag{12}$$

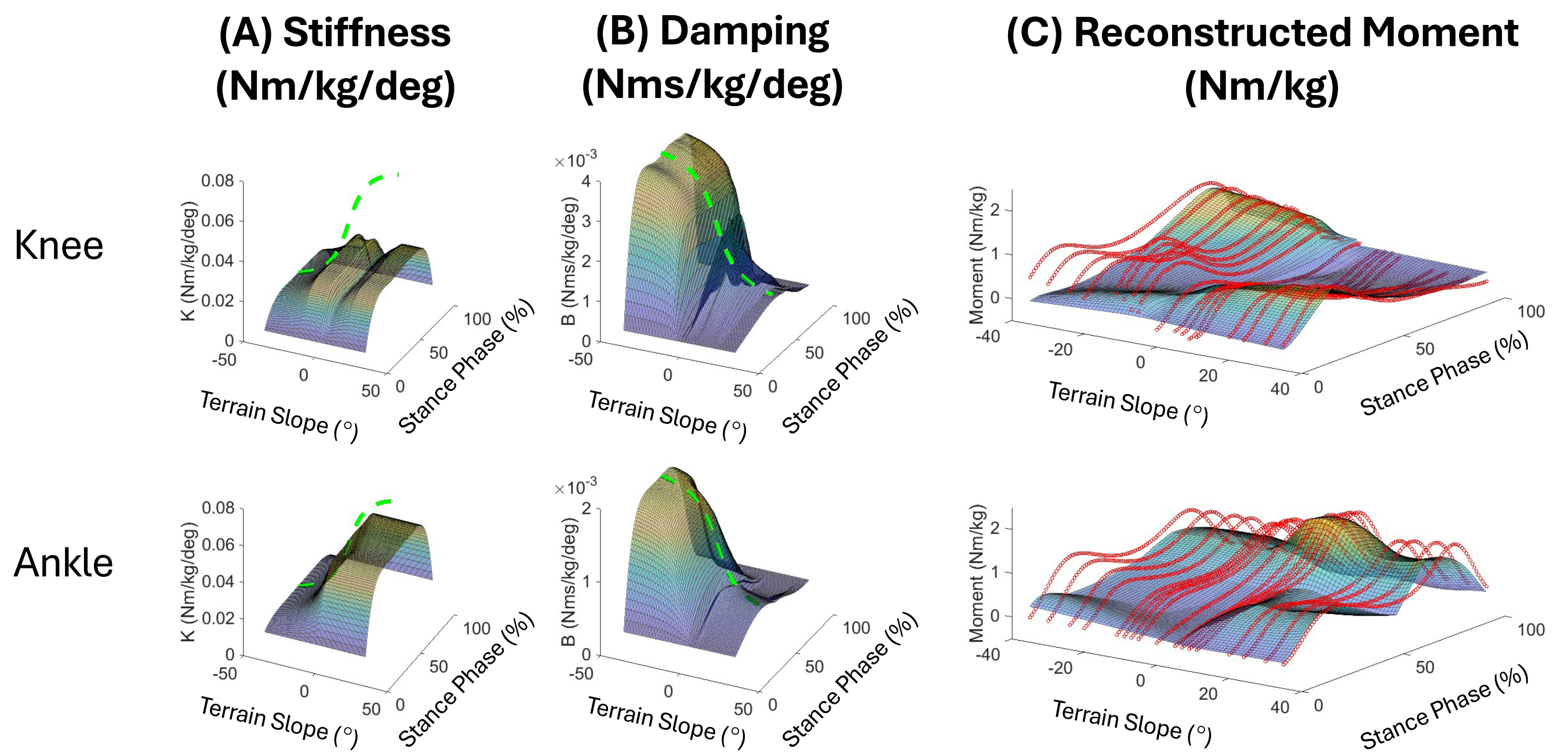

**Fig. 7.** Continuous impedance parameterization. **(A)** Identified stiffness and **(B)** damping surfaces. Green dashed curves denote slope-dependent upper bounds $K_{max}(s)$ and $B_{max}(s)$. **(C)** Moments reconstructed from the impedance model closely match the AB reference (red dotted lines).

The equality constraints prescribed minimum impedance values at heel contact and maintained the terminal ES impedance values during the initial LS support interval. Additional inequality constraints enforced the prescribed impedance bounds and regulated the progression of stiffness and damping. Specifically, derivative-based shape constraints promoted increasing impedance during initial loading, approximately constant parameters during single support, and constant or decreasing values during unloading. The complete regression-matrix construction and equality, bound, and shape constraints are provided in Appendix B.

The identified impedance parameters were validated by reconstructing the target joint moments. Figure 7 illustrates the resulting stiffness and damping surfaces together with the reconstructed moments, while quantitative reconstruction performance is summarized in Table 1. For the LW and ascent models, the reconstructed moments closely matched the target AB trajectories, yielding high overall squared Pearson correlation coefficient $r^2$ values for both the knee (0.94 ± 0.04) and ankle (0.89 ± 0.03). High ES and LS accuracies indicate that the proposed parameterization preserves the primary biomechanical characteristics of the target moments. The descending models exhibited lower overall $r^2$ values, primarily due to reduced ES reconstruction accuracy. This behavior was expected because the proposed $\theta_{sup}$-based yielding strategy intentionally does not reproduce the initial stance phase moment peak during weight acceptance. As a result, the knee and ankle ES models achieved $r^2$ values of 0.32 ± 0.36 and 0.13 ± 0.08, respectively. In contrast, the LS models maintained substantially higher accuracies (0.73 ± 0.21 for the knee and 0.91 ± 0.03 for the ankle), demonstrating that the proposed parameterization accurately reproduces the support and unloading dynamics prioritized by the controller despite the reduction in early stance reconstruction performance.

**Table 1.** Phase-aligned $r^2$ values of reconstructed moment models.

| | LW + Ascent Overall | LW + Ascent ES | LW + Ascent LS |
|---|---|---|---|
| **Knee** | $0.94 \pm 0.04$ | $0.98 \pm 0.01$ | $0.83 \pm 0.15$ |
| **Ankle** | $0.89 \pm 0.03$ | $0.67 \pm 0.19$ | $0.84 \pm 0.10$ |
| | **Descent Overall** | **Descent ES** | **Descent LS** |
| **Knee** | $0.28 \pm 0.25$ | $0.32 \pm 0.36$ | $0.73 \pm 0.21$ |
| **Ankle** | $0.48 \pm 0.29$ | $0.13 \pm 0.08$ | $0.91 \pm 0.03$ |

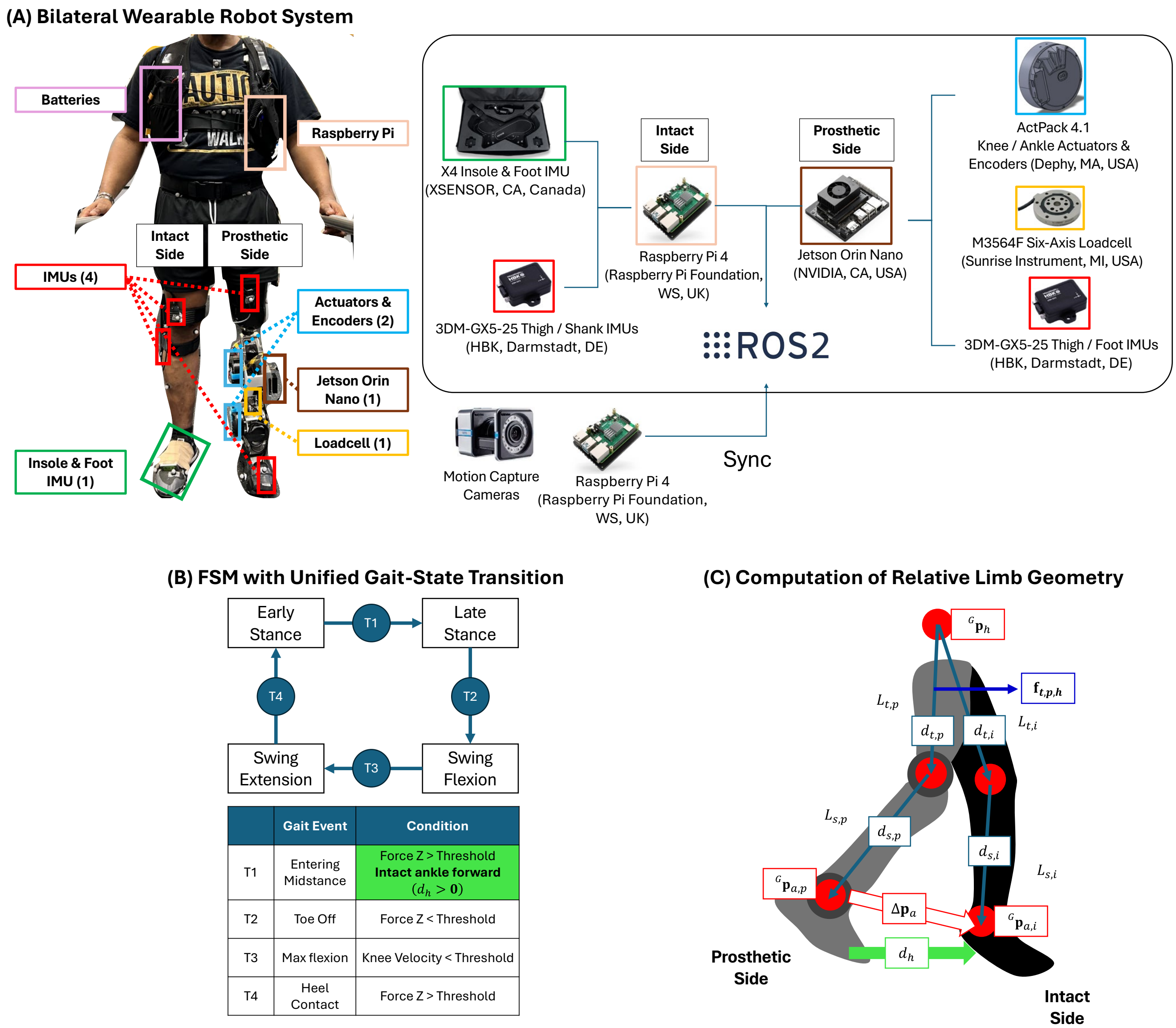

| | Gait Event | Condition |
|---|---|---|
| T1 | Entering Midstance | Force Z > Threshold **Intact ankle forward** $(d_h > 0)$ |
| T2 | Toe Off | Force Z < Threshold |
| T3 | Max flexion | Knee Velocity < Threshold |
| T4 | Heel Contact | Force Z > Threshold |

**Fig. 8.** Experimental setups are illustrated. **(A)** The proposed wearable system incorporates prosthetic and intact side sensing system to capture both limb dynamics simultaneously. **(B)** The proposed FSM was enhanced with the relative geometry of limbs, enabling slope-independent T1 condition. **(C)** The relative geometry between both ankles $d_h$ was computed through forward kinematics using the bilateral sensing system.

## IV. Experimental Methods

### A. Prosthetic System with Bilateral Sensing

The proposed controller was implemented on the second-generation Open-Source Leg (OSL), a powered knee-ankle prosthesis platform developed by the University of Michigan (Azocar *et al.*, 2020; Best *et al.*, 2025). The system incorporated bilateral sensing from both limbs (Hu, Rouse and Hargrove, 2018; Su *et al.*, 2019; Posh and Gregg, 2025), onboard computation, and impedance-controlled actuation for continuous terrain-adaptive locomotion, as illustrated in Fig. 8A.

The prosthesis incorporated powered knee and ankle actuators (Actpack 4.1, Dephy, MA, USA) with embedded joint encoders and a six-axis load cell (M3564F, Sunrise Instrument, MI, USA) mounted between the knee and ankle modules. Four 9-axis IMUs (3DM-GX5-25, HBK, Darmstadt, Germany) were mounted on the prosthetic-side thigh and foot and the intact-side thigh and shank. The intact limb was additionally instrumented with an XSENSOR X4 module (X4, XSENSOR, Calgary, Canada), comprising a foot-mounted IMU and a pressure-sensitive insole. All onboard sensing and control algorithms were executed on a Jetson Orin Nano under ROS 2 Jazzy at 100 Hz. Motion capture measurements were synchronized with the ROS 2 data stream for post-hoc analysis.

### B. Controller Implementation

Our mode-unified framework employs a four-state FSM that is shared across all locomotion modes (Fig. 8B), similar to other studies (Sup, Bohara and Goldfarb, 2008; Simon *et al.*, 2014; Bhakta *et al.*, 2020; Camargo *et al.*, 2023). Transition conditions (T1-T4) are also summarized in Fig. 8B. T1 was defined from bilateral limb progression, corresponding to the instant when the intact foot advances beyond the prosthetic foot during stance with prosthetic load beyond a threshold. The remaining transitions (T2-T4) followed conventional prosthesis control logic.

The intact and prosthetic limb geometries were reconstructed from wearable IMU orientations and subject-specific segment lengths, as illustrated in Fig. 8C. Let ${}^{G}\mathbf{p}_h$ denote the hip joint position in the global frame. The segment direction vectors $d_{t,i}$, $d_{s,i}$, $d_{t,p}$, and $d_{s,p}$, together with the corresponding segment lengths $L_{t,i}$, $L_{s,i}$, $L_{t,p}$, and $L_{s,p}$ were used to reconstruct the intact and prosthetic ankle positions ${}^{G}\mathbf{p}_{a,i}$ and ${}^{G}\mathbf{p}_{a,p}$ through the forward kinematic relationships,

$$\begin{aligned} {}^{G}\mathbf{p}_{a,i} &= {}^{G}\mathbf{p}_h + L_{t,i}d_{t,i} + L_{s,i}d_{s,i} \\ {}^{G}\mathbf{p}_{a,p} &= {}^{G}\mathbf{p}_h + L_{t,p}d_{t,p} + L_{s,p}d_{s,p} \end{aligned} \tag{13}$$

The relative ankle position was then computed as

$$\Delta\mathbf{p}_a = {}^{G}\mathbf{p}_{a,i} - {}^{G}\mathbf{p}_{a,p} \tag{14}$$

The prosthetic forward direction $\mathbf{f}_{t,p,h}$ was obtained by projecting the prosthetic thigh IMU local positive $z$-axis onto the horizontal plane and normalizing the resulting vector. The limb-crossing metric was subsequently computed by projecting the inter-ankle vector onto this forward direction,

$$d_h = \Delta\mathbf{p}_a^T\mathbf{f}_{t,p,h} \tag{15}$$

where $d_h \geq 0$ indicate that the intact ankle has advanced beyond the prosthetic ankle in the prosthesis-referenced walking direction and was used to trigger the T1 transition.

Within the unified FSM, the slope-dependent impedance parameters derived in Section III were executed using the load-reactive impedance strategy proposed by Simon *et al.* (Simon *et al.*, 2014). During level walking and ascending locomotion, stance phase stiffness and damping were continuously interpolated according to the normalized prosthetic side vertical load, while the equilibrium angle followed the prescribed trajectory shown in Fig. 5B. During descending locomotion, the same load-reactive interpolation was applied to stiffness and damping, whereas the equilibrium angle followed the proportional yielding strategy described in Section III. During swing flexion and extension, constant stiffness and damping values were used, while the equilibrium angles were assigned to the slope-dependent maximum swing-flexion and heel-contact angles, respectively. To improve robustness against variations in user loading and gait-state transition timing, the normalized vertical load at the ES-to-LS transition was stored and subsequently used as the reference during the unloading phase. A representative implementation of the unified controller is illustrated in Fig. A2 (Appendix C), showing the synchronized evolution of the FSM states, prosthetic side vertical load, and continuously varying impedance parameters throughout the gait cycle.

### *C. Experimental Protocol*

Ten individuals with unilateral transfemoral amputation (eight male and two female; 5 right and 5 left amputations or limb difference; age: 39.7 ± 11 years; body mass: 85.9 ± 13.9 kg; height: 1.74 ± 0.13 m; years since amputation: 17.9 ± 11.7 years) participated in this study. All participants provided written informed consent in accordance with the approval by the Georgia Institute of Technology Institutional Review Board (H21117).

Before testing, the prosthesis was fit and aligned by a certified prosthetist during quiet standing. Participants then completed a static calibration trial to initialize the wearable IMUs and establish a common global reference frame, followed by an acclimation period consisting of level walking (0 °), ramps (± 9 °), and stairs (± 24 °). After acclimation, swing-phase stiffness and damping were adjusted by up to 10% from their nominal values to achieve comfortable limb advancement. In addition, constant heel-contact angle offsets of up to 10° were introduced during descending locomotion to promote a more extended knee and reduced ankle plantarflexion at initial contact, improving heel-contact stability while preserving the continuous slope-dependent trajectories. This adjustment was introduced because direct reproduction of the AB heel-contact resulted in excessive knee flexion and ankle plantarflexion which was destabilizing in prosthesis users.

The proposed controller was subsequently evaluated during level walking, ramp walking, and stair ambulation, as illustrated in Fig. 1C. Level walking and ramps (0 ° - 13 °) were performed on an instrumented incline/decline treadmill at 0.4 - 0.5 m/s, whereas stair ascent and descent were evaluated over slopes ranging from 15 ° to 35 °. The intermediate 13 ° ramp and 15 ° stair conditions were not represented in the underlying AB biomechanics dataset and were included to evaluate controller performance in terrain regions absent from the reference data. Due to experimental time constraints, the 13 ° ramp and 15 ° stair conditions were completed by 9 and 6 of the 10 participants, respectively. Participant inclusion in these conditions was determined by available testing time rather than K-level or task performance. Whole-body kinematics and ground reaction forces were recorded using a motion-capture system synchronized with an instrumented treadmill and stair force plates for post-hoc analysis. During stair ascent, participants initiated the first step with the intact limb, whereas the prosthetic limb served as the leading limb during descent in accordance with how stair negotiation is commonly trained in clinical practice.

### *D. Statistical Analysis*

The weighted squared Pearson correlation coefficient ($r^2$) was used to quantify how closely the experimentally measured prosthetic outcomes followed the slope-dependent trends of the corresponding model predictions and AB reference data. To account for the non-uniform spacing of terrain slopes, each slope was weighed according to the interval it represented within the slope domain, based on the spacing to adjacent slope values.

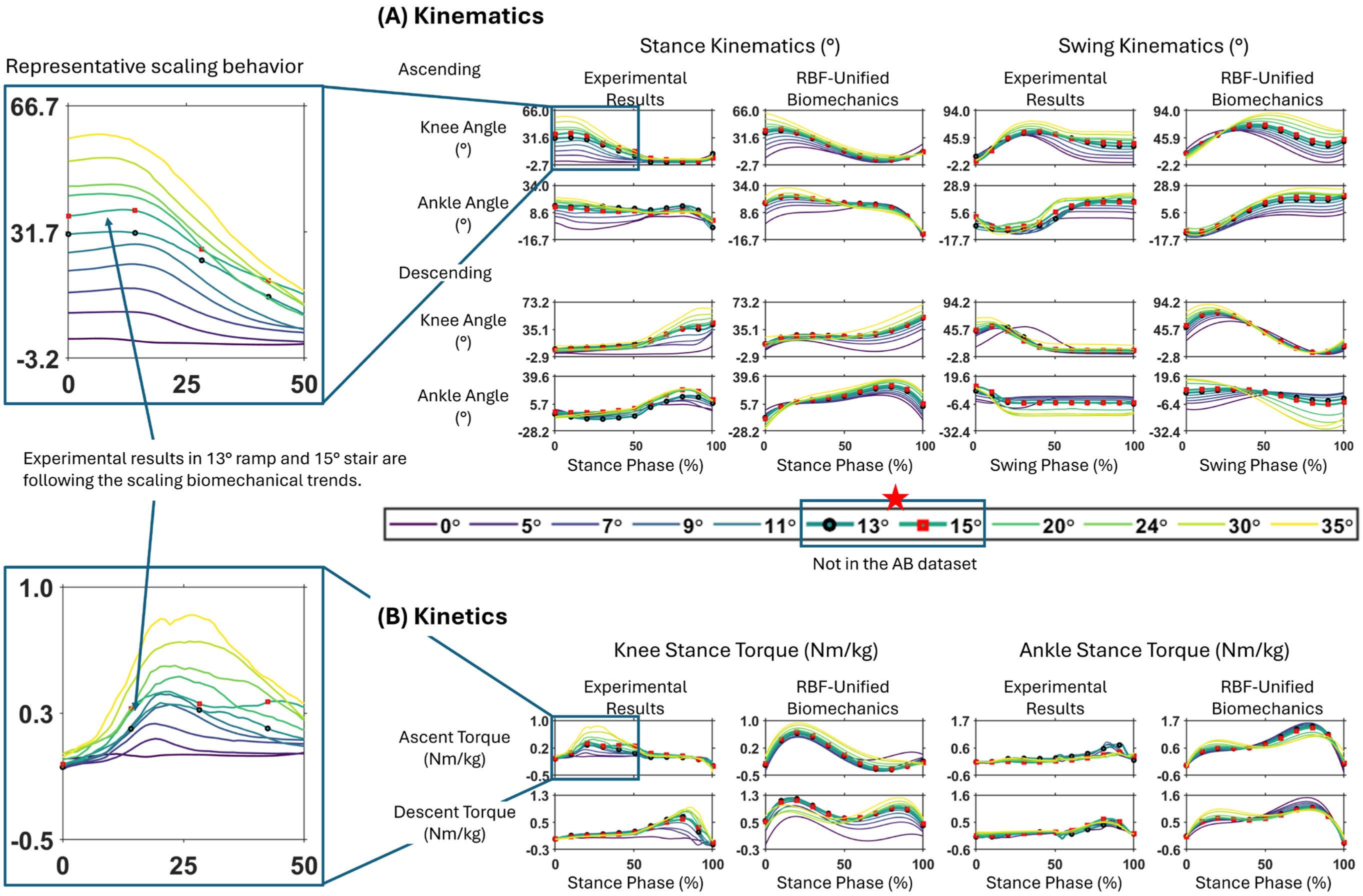

**Fig. 9.** Experimental results on inter-participant average (n = 10) **(A)** kinematics and **(B)** kinetics are illustrated, by comparing with the RBF-unified biomechanics models derived in Section II. The enlarged panels on the left highlight representative early stance knee angle and torque to illustrate the observed slope-dependent scaling. Red star (★) in the legend indicates the intermediate 13 ° ramp (lines with circles, n = 9) and 15 ° stair (lines with squares, n = 6) conditions, which were not included in the AB dataset used to construct the partitioned polynomial models.

## V. Results

The experimentally measured prosthesis kinematics and kinetics are compared with the target RBF-unified biomechanics across the continuous terrain slope domain in Fig. 9. Notably, the proposed controller generated smooth biomechanical trajectories at the intermediate 13° ramp and 15° stair conditions (★), although neither terrain was represented in the underlying AB biomechanics dataset. The measured trajectories remained consistent with the continuous slope progression predicted by the RBF-unified model, demonstrating successful interpolation to terrain regions absent from the reference dataset.

Overall, the proposed controller reproduced the intended slope-dependent joint behaviors across both the knee and ankle. During ascending locomotion, knee flexion and ankle dorsiflexion increased progressively with terrain slope throughout stance, while larger swing-phase flexion and clearance angles were observed on steeper terrains. During descending locomotion, the controller generated progressively greater knee yielding together with increased ankle dorsiflexion during late stance. Although subject-dependent timing differences were present, the experimental trajectories preserved the continuous slope-dependent progression predicted by the RBF-unified biomechanics.

An additional quantitative evaluation was performed using representative kinematic landmarks and peak moments (Fig. 10). The experimentally measured quantities closely followed both the RBF-unified model and the underlying AB reference across the terrain domain. During ascent, heel-contact knee flexion, heel-contact ankle dorsiflexion, swing clearance angles, and peak knee support moments increased monotonically with terrain slope. During descent, increasing terrain steepness produced progressively greater knee toe-off angles and yielding behavior while maintaining relatively constant ankle toe-off, consistent with the proportional-yielding strategy introduced in Section III. For the ankle, the controller also preserved the continuous slope-dependent progression of heel-contact dorsiflexion, swing clearance, and toe-off angle, demonstrating that the unified parameterization successfully reproduced the intended kinematic adaptation across both ascending and descending locomotion.

Figs. 10A and 10B summarize the slope-dependent scaling performance using $r^2$ for each kinematic landmark and peak torque. Here, kinematic landmarks refer to joint angles at specific gait events, including heel contact during both ascending and descending locomotion, swing clearance during ascent, and toe-off during descent. Knee-related metrics showed high agreement between the proposed unified controller and the RBF-unified model ($r^2$ = 0.82 - 0.99), while ankle kinematic metrics ranged from 0.78 to 0.96. In contrast, ankle stance moment peaks showed rather low agreement ($r^2$ = 0.08 during ascent and 0.25 during descent). High agreement was also observed between the unified controller and the AB reference, with knee-related metrics ranging from 0.75 to 0.99 and ankle kinematic metrics from 0.85 to 0.97, indicating that the proposed controller successfully reproduced the AB biomechanical characteristics encoded by the unified model. The ankle stance torque peaks likewise exhibited a medium level of agreement with the AB reference ($r^2$ = 0.47 during ascent and 0.51 during descent).

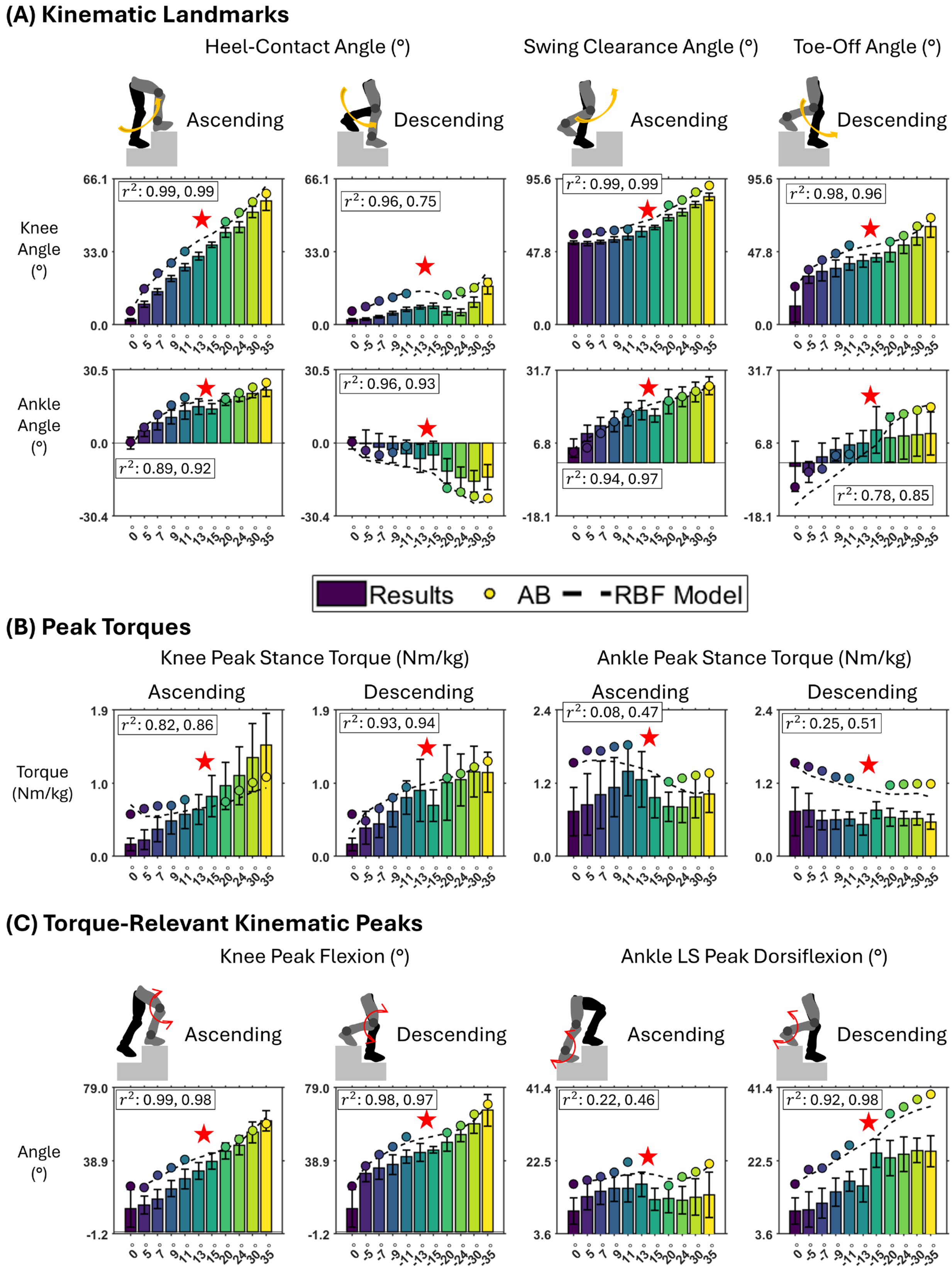

**Fig. 10.** Representative **(A)** kinematic landmarks **(B)** peak torques, and **(C)** torque-relevant kinematic peaks across terrain slope. Experimental results from the proposed unified controller are compared with the RBF-unified biomechanics model and the underlying AB reference. The left and right $r^2$ values in each subplot denote squared Pearson correlation coefficients between the experimental results and the RBF-unified model and AB data, respectively. Bars represent mean ± 1 SD. Red star (★) in each subplot indicates the intermediate 13 ° ramp and 15 ° stair conditions, which were not included in the AB biomechanics dataset. In the posture schematics, orange arrows in **(A)** indicate prosthetic limb progression, whereas red arrows in **(C)** indicate the direction of joint torque.

Further inspection of the torque-relevant kinematic peaks in Fig. 10C was conducted to examine how the realized joint kinematics contributed to the stance-torque outcomes. Under the impedance relationship in Eq. (7), the stiffness-generated joint torque depends on the angular displacement from the modeled equilibrium angle $(\theta - \theta_{eq})$; therefore, insufficient realization of the corresponding joint excursion can reduce the achievable stance torque. Averaged across terrain slopes, the experimentally realized knee peak stance torques reached 101.4% and 79.3% of the RBF-model values during ascent and descent, respectively, while peak knee flexion reached 80.0% and 85.6% of the corresponding model values. In comparison, ankle peak stance torques reached only 74.8% and 56.2% of the RBF-model values during ascent and descent, respectively, accompanied by late-stance (LS) peak dorsiflexion reaching only 73.7% and 66.2% of the model values. Thus, the greater under-realization of ankle torque coincided with insufficient LS dorsiflexion relative to the RBF-unified biomechanics, providing a mechanical explanation for the lower ankle stance-torque agreement observed in Fig. 10B.

## VI. Discussion

The primary contribution of this work is the demonstration that lower-limb prosthesis control can be formulated using a continuous biomechanical representation spanning level walking, ramps, and stairs, rather than as a collection of independently designed locomotion modes. Existing biomimetic prosthesis controllers generally derive impedance parameters from discrete AB locomotion datasets and therefore inherit the same mode-specific structure, requiring separate controller designs and transition logic for each ambulation task. In contrast, the proposed framework represents joint biomechanics as continuous functions of terrain slope and gait phase, enabling both biomechanical modeling and controller parameterization within a unified domain. This formulation shifts controller design from defining discrete locomotion modes to parameterizing a continuous environmental variable, providing a more scalable foundation for future terrain-adaptive prosthesis control and real-world mobility. The results support two hypotheses introduced in Section I. First, the RBF-unified biomechanics preserved the reconstruction accuracy of the locally optimized ascent and descent models while outperforming the global polynomial representation, supporting **H1**. Second, experiments with ten individuals with transfemoral amputation demonstrated that the unified controller reproduced the principal slope-dependent biomechanical trends of healthy individuals, including intermediate terrain conditions absent from the reference dataset, supporting **H2**. However, **H2** was only partially supported for ankle stance kinetics, where the experimental torque trends exhibited lower agreement both with the unified model and AB reference.

The proposed continuous formulation provides two advantages. First, it enables direct interpolation to intermediate terrain conditions not present in the reference dataset, as demonstrated experimentally for the 13 ° ramp and 15 ° stair conditions. Second, because the controller is parameterized continuously with respect to terrain slope, a single controller architecture can span the entire modeled terrain domain without introducing separate mode-specific parameter sets or additional FSM-based transitions. This reduces controller complexity while preserving continuous scaling in biomechanical behavior across the modeled terrain domain.

The proposed RBF-unified biomechanics addresses a fundamental limitation of polynomial-based locomotion models by reconciling local biomechanical fidelity with global continuity. While partitioned polynomial models accurately reconstructed each locomotion subset, directly concatenating them introduced discontinuities near the shared level-walking condition, whereas a single global polynomial excessively smoothed the local biomechanical characteristics. The RBF blending preserved the locally optimized biomechanics while providing a continuous representation suitable for real-time controller parameterization. Importantly, the objective of the continuous biomechanical representation is not to reconstruct unknown AB biomechanics exactly, but to provide a continuous biomimetic prior for controller design. This distinction is particularly relevant within the transition region between steep ramps and shallow stairs, where changes in foot-ground contact geometry may produce substantially different ankle biomechanics despite similar terrain inclinations. Therefore, the successful implementation at the 13 ° ramp and 15 ° stair conditions should be interpreted as validation of the continuous controller parameterization rather than confirmation of the interpolated AB biomechanics themselves.

Although unified biomechanics provide continuous reference trajectories, directly reproducing AB joint kinematics and kinetics is neither feasible nor necessarily desirable for powered prostheses. The unified biomechanics served as a design prior rather than a direct control objective, requiring transformation into impedance parameters that balance biomimetic behavior with practical implementation. Rather than optimizing all three impedance parameters simultaneously (Best *et al.*, 2023; Cortino, Best and Gregg, 2024; Hong, Naseri and Huang, 2026), the proposed framework derived stiffness and damping after designing equilibrium angles (Fig. 5B) based on the bio-inspired support angle $(\theta_{sup})$ definition (Fig. 4). The $\theta_{sup}$ formulation, constrained optimization, and proportional yielding strategy (Simon *et al.*, 2014; Bhakta *et al.*, 2020; Camargo *et al.*, 2023) collectively reduced the controller dimensionality while preserving the biomechanical behaviors most relevant to prosthesis function, including limb support, controlled yielding, and swing clearance.

Interestingly, the constrained LSQ optimization consistently converged to trapezoid-like stiffness and damping profiles throughout stance (Figs. 7A and 7B). These profiles emerged naturally from the optimization while closely following the prescribed arctangent upper-bound envelopes, indicating that the constraints actively regulated the resulting impedance parameters. An exception was observed for knee stiffness during ascending locomotion, where the optimized values remained below the allowable upper bound despite accurate moment reconstruction (Figs. 7A and 7C). This behavior suggests that slope-dependent knee support was achieved primarily through modulation of the equilibrium angle trajectory rather than increasing joint stiffness, highlighting the complementary roles of stiffness and equilibrium angle within the proposed parameterization.

The transformation from AB biomechanics to prosthesis impedance control required deliberate departures from reference joint dynamics. This was particularly important during descending locomotion, where the initial knee moment peak obtained from inverse dynamics represents the net mechanical moment at the biological joint rather than a direct actuator torque requirement. Such net moments reflect the combined effects of active muscle forces and passive musculoskeletal structures (Zajac, Neptune and Kautz, 2002) and therefore cannot be interpreted as one-to-one motor commands. Specifically, during stair descent, the biological knee extensors dissipate energy eccentrically while the knee flexes to lower the body (Cluff and Robertson, 2011). Therefore, directly reproducing the initial moment peak with a large prosthetic support torque would oppose the controlled knee yielding required to "ride" the prosthesis during descent (Simon *et al.*, 2014). Beyond this departure from the AB knee moment during descent, several additional design choices intentionally prioritized prosthesis functionality over one-to-one reproduction of natural biomechanics. The ankle support equilibrium angle was bounded to 10 ° dorsiflexion to remain within the prosthesis operating range and preserve sufficient plantarflexion torque. Stance biomechanics were represented through a three-region equilibrium angle structure, with stiffness and damping constrained to physically realizable bounds and progression patterns rather than directly reproducing the full AB moment trajectory. During swing, the controller similarly targeted functional landmarks (e.g. maximum flexion for limb clearance and heel-contact angle for preparation for loading) rather than tracking the complete continuous AB kinematic trajectory. Collectively, these design choices preserved the biomechanical functions considered most relevant to prosthesis control while intentionally relaxing exact replication of natural joint behavior.

Remaining discrepancies primarily reflect implementation choices rather than limitations of the underlying biomechanics. The largest kinematic differences occurred at heel-contact for both joints, consistent with the constant heel-contact angle offsets introduced to improve prosthetic stability during initial contact (Section IV-C). The ankle kinetic discrepancy highlights a tradeoff introduced by the mode-unified state transition strategy. In conventional mode-specific control, the ES-to-LS transition can incorporate an ankle angle threshold, ensuring that sufficient dorsiflexion develops before entering late stance and thereby preserving the angular displacement $(\theta - \theta_{eq})$ required for modeled torque. Such thresholds, however, vary substantially across different modes and are difficult to retain within a single mode-independent transition rule. The proposed bilateral limb-progression criterion removes this dependence but decouples the ES-to-LS transition from the desired peak dorsiflexion angle. Consequently, LS may be initiated before sufficient dorsiflexion develops, limiting the angular displacement and the resulting ankle torque. This consideration is particularly relevant during steep stair descent, where the RBF-unified ankle dorsiflexion trajectory can exceed the mechanical range of the current prosthesis (Fig. 10C, Ankle LS Peak Dorsiflexion for -20 °, -24 °, -30 °, and -35 °), making direct reproduction of the reference kinematics infeasible. Accordingly, strategies such as prescribing a more plantarflexed equilibrium angle or accelerating the progression of impedance parameters may help generate greater stance ankle torque without requiring full reproduction of the modeled dorsiflexion. Such modifications may be particularly important during late stance, consistent with prior findings highlighting the biomechanical importance of push-off timing and power generation (Malcolm *et al.*, 2015; Zelik and Adamczyk, 2016; Fylstra *et al.*, 2020).

A further difference was observed during the swing extension, where the experimental trajectories exhibited plateau regions immediately before initial contact (Fig. 10A). This behavior is consistent with terminal-swing stabilization strategies previously reported in unilateral prosthesis users. Prior studies have identified a prolonged latency between full prosthetic knee extension and initial contact, interpreted as a period during which users ensure a stable position before loading (Mâaref *et al.*, 2010), and have reported corresponding adaptations in terminal-swing muscle activation (Wentink *et al.*, 2013). More recent work further demonstrated that when prosthetic knee extension was completed earlier, users tended to preserve or increase this latency rather than initiate ground contact earlier (Marsh, Haake and Goldfarb, 2026). Thus, the observed plateau may reflect user-selected preparation for stable initial contact rather than a limitation of the continuous biomechanical representation.

Beyond the impedance parameter formulation, successful controller implementation also relied on the proposed bilateral sensing architecture (Fig. 8). Compared with conventional mode-specific controllers, this architecture eliminates mode-dependent gait state definitions while preserving a single FSM across all modeled terrain conditions. The relative limb geometry was estimated using a simplified sagittal-plane forward kinematics model based on wearable IMUs, without explicitly accounting for out-of-plane pelvic motion or hip tilt. Consequently, the computed geometry should be interpreted as a robust gait state variable rather than an anatomically exact estimate of bilateral limb configuration. Despite this simplification, the experimental results demonstrated reliable gait state transitions across all evaluated terrain conditions, suggesting that the proposed geometry provides sufficient information for unified controller implementation.

For additional reference, the biomechanical replication results of the proposed controller were examined alongside previously reported results using a mode-specific, hand-tuned controller (Zhou *et al.*, 2025), as detailed in Appendix and Fig. 11. Although both studies evaluated common terrain conditions (± 5 ° ramps and ± 24 ° stairs), they involved different participant groups, experimental protocols, and hardware capabilities. In specific, the previous study employed the first-generation OSL (Azocar *et al.*, 2020), with ± 15 ° of ankle motion, whereas the proposed study used the second-generation OSL (Best *et al.*, 2025), with an expanded range of ± 30 °. This hardware difference is particularly relevant to stair locomotion, where the AB-referenced ankle heel-contact and swing clearance angles exceeded the physical limit of the earlier OSL. Therefore, the observed differences in biomechanical replication cannot be attributed solely to the control strategies, and the comparison should be interpreted as descriptive rather than a direct assessment of controller superiority. Detailed comparison methods and numerical results are provided in Appendix D.

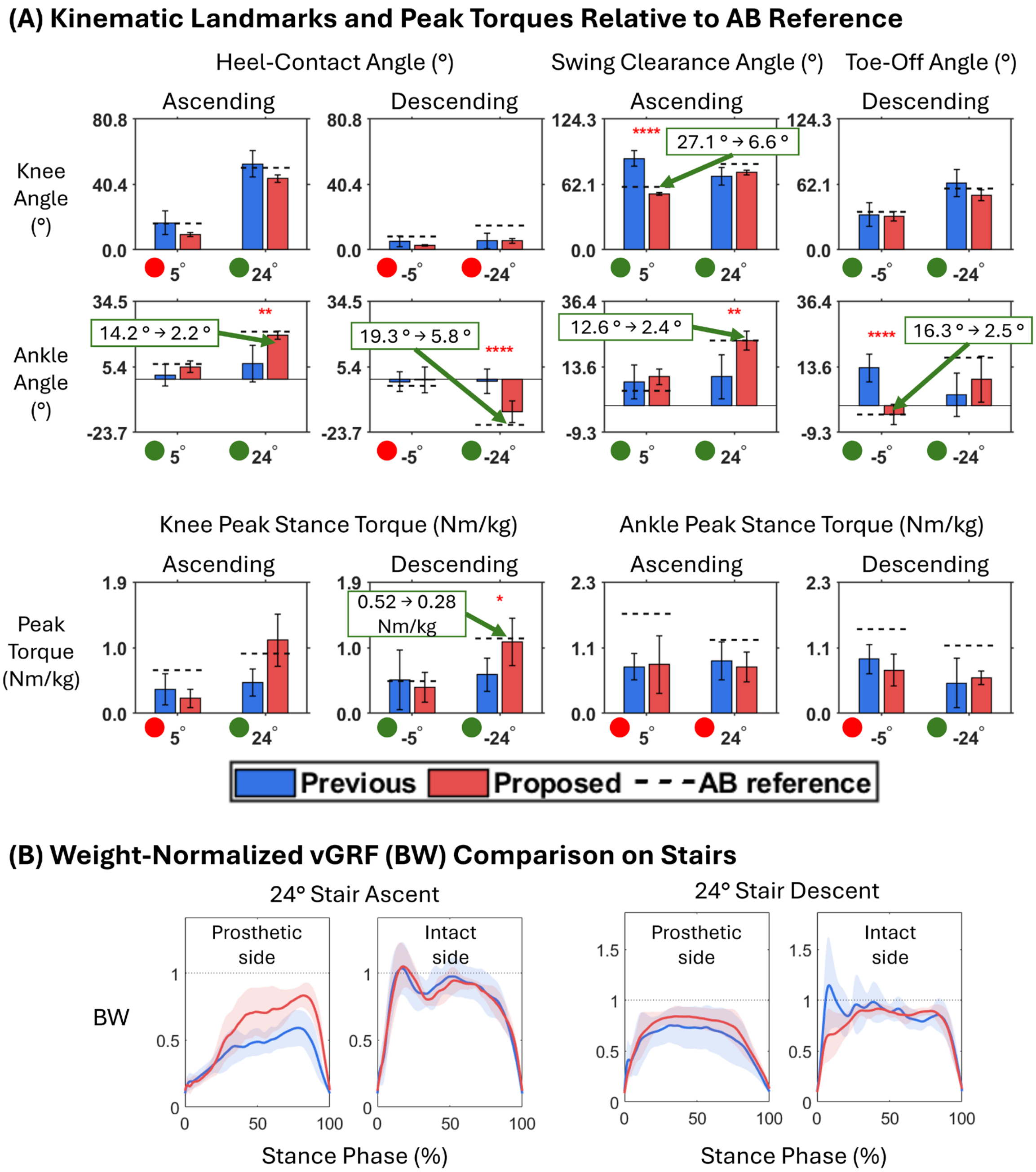

**Fig. 11.** Descriptive comparison of biomechanical outcomes from the previously reported mode-specific controller (Zhou et al., 2025) and the proposed mode-unified controller relative to the AB reference. **(A)** Knee and ankle kinematic landmarks and peak stance torques across common terrain conditions (± 5 ° ramps and ± 24 ° stairs). Bars and error bars denote cross-subject mean ± 1 SD for the previous study (blue, n=9) and the proposed unified controller (red, n = 10), while dashed black lines indicate the corresponding AB reference values. Green circles indicate lower MAEs with the proposed controller, whereas red circles indicate no improvement. Green-boxed annotations and arrows highlight the six statistically significant between-study differences favoring the proposed controller, along with the corresponding MAE reductions relative to the AB reference. Significance was assessed using Welch's two-sample t-tests (*$p < 0.05$, **$p < 0.01$, ***$p < 0.001$, ****$p < 0.0001$). **(B)** Weight- and stance phase-normalized vertical ground reaction force (vGRF) profiles of the prosthetic and intact sides during 24 ° stair ascent and descent. Solid lines and shaded regions denote cross-subject mean ± 1 SD.

Nevertheless, the proposed framework offers a distinct advantage in controller parameterization. A conventional two-joint, four-state impedance FSM requires at least 24 independently specified stiffness, damping, and equilibrium-angle parameters per locomotion mode (Zhou *et al.*, 2025), (Simon *et al.*, 2014). In contrast, the proposed framework derives these parameters continuously from terrain slope and gait phase using a unified biomechanical representation, without independently hand-tuning a complete parameter set for each mode. This formulation reduces the reliance on mode-specific controller design while retaining the principal slope-dependent biomechanical characteristics demonstrated in proposed framework.

Beyond the AB-replication analysis, Fig. 11B provides additional insight into how participants loaded the prosthesis during the matched overground stair conditions. During stair ascent, the proposed controller was associated with greater prosthetic-side vGRF throughout stance while intact-side loading remained comparatively similar between controller groups. This pattern suggests greater weight acceptance through the prosthetic limb rather than a compensatory increase in intact-side loading. Stair descent loading profiles were comparatively similar between controllers. Because the present analysis was limited to stance-phase vGRF profiles and did not include complementary measures such as vertical impulse or loading symmetry, these findings should be interpreted as evidence of altered limb-loading behavior rather than a direct measure of functional improvement (Cutti *et al.*, 2018; Kobayashi *et al.*, 2022).

The present study has several limitations that motivate future research. First, the proposed controller assumed continuous terrain slope information as a high-level input. Although the controller architecture is independent of how this information is obtained, integrating onboard slope estimation methods based on wearable sensing (Best *et al.*, 2023; Lee, Lee and Young, 2024; Kim *et al.*, 2025) or environment perception (Gao *et al.*, 2020; Cheng *et al.*, 2025; Anselmino, Simon and Hargrove, 2026) will be necessary for fully autonomous operation.

Second, the proposed continuous biomechanics were derived from AB datasets collected under self-selected or relatively slow walking speeds (up to 0.6 m/s). Walking speed was not explicitly included as a model parameter, and the resulting impedance representation is expected to be most applicable within similar speed ranges. Extending the framework to incorporate continuous speed adaptation represents an important direction for future work and could enable unified control across both terrain and walking speed variations (Fey *et al.*, 2014; Lenzi, Hargrove and Sensinger, 2014; Best *et al.*, 2023).

Third, the unified biomechanics were constructed by blending independently modeled ramp and stair datasets. The interpolated biomechanics within the transition region between steep ramps and shallow stairs should be interpreted as controller parameterization rather than true AB biomechanics.

Despite these limitations, the proposed framework establishes a continuous biomimetic foundation for terrain-adaptive prosthesis control that unifies biomechanical modeling, impedance parameterization, and controller implementation within a single representation. Future integration of autonomous terrain estimation, continuous speed adaptation, and user-specific optimization has the potential to further extend this framework toward practical daily-life prosthesis control.

## VII. Conclusion

This paper presented a mode-unified biomimetic control framework that represents lower-limb biomechanics continuously as functions of terrain slope and gait phase. The RBF-unified models preserved the accuracy of locally optimized biomechanics while enabling continuous impedance parameterization across level walking, ramps, and stairs. Experiments with ten prosthesis users demonstrated continuous terrain adaptation, including 13 ° ramp and 15 ° stair conditions absent from the reference AB dataset. The proposed controller reproduced the principal slope-dependent trends ($r^2$ = 0.82 - 0.99 for all knee-related measures and 0.78 - 0.96 for ankle kinematics), while ankle stance torque remained less accurately reproduced. Together, these findings support our unified biomimetic control framework that reduces mode-specific tuning while enabling users to address varied terrain without relying on separately parameterized controllers for each locomotion modes.

## Author Note

During the preparation of this work, the authors used Grammarly and OpenAI ChatGPT (GPT-5.6 Sol) to assist with language refinement. ChatGPT was also used for MATLAB code review and polishing and refinement of data-visualization scripts. All AI-assisted code, visualizations, and manuscript text were reviewed, verified, and edited by the authors.

## Acknowledgements

The authors sincerely thank all participants who took part in this study. The authors also acknowledge the EPIC Prosthesis VIP Team for their assistance with experiments and data processing. Special thanks are extended to Chase Sun for his substantial technical support in mechatronics development.

## Ethical Considerations

The experimental protocol was approved by the Georgia Institute of Technology Institutional Review Board (H21117). All participants provided written informed consent to participate in the study and for the publication of their data, including images and videos.

## Funding

The authors disclosed receipt of the following financial support for the research, authorship, and/or publication of this article: This work was supported by the U.S. Department of Defense Congressionally Directed Medical Research Programs (CDMRP) under Award No. W81XWH-22-1-1091 and the National Science Foundation (NSF) Sustainable Development of Smart Medical Devices (SUSMED) Fellowship.

## Declaration of Conflicting Interests

The authors declared no potential conflicts of interest with respect to the research, authorship, and/or publication of this article.

## SUPPLEMENTAL MATERIAL

Supplementary multimedia material demonstrating the proposed mode-unified prosthesis controller across different locomotion conditions is publicly available on YouTube. MATLAB code and related resources for the unified biomechanics modeling and impedance parameterization framework are provided through the authors' GitHub repositories.

## APPENDIX

### A. Formulation of RBF-unified Biomechanics

Let $f_a(s_a, p)$ and $f_d(s_d, p) \in \mathbb{R}^{101\times36}$ denote the local full-ascent and full-descent models obtained from Section II-B, respectively. Samples generated from both models over the full slope-phase domain were treated as observations in a two-dimensional space defined by terrain slopes $s_a \in \{0, 1, \ldots, 35\}$, $s_d \in \{-35, -34, \ldots, -1, 0\}$, and $p \in P$, as defined in Eq. (3). The samples from both partitions were stacked to

$$M = [f_a \; f_d] \in \mathbb{R}^{101\times72} \tag{A1}$$

This stacked representation includes the level-walking condition from both partitions, resulting in duplicated samples at $s = 0\,°$. The objective of RBF interpolation is to derive a unified representation of biomechanics $f_U(s_U, p) \in \mathbb{R}^{71\times101}$ for each joint in the slope-phase domain, where $s_U \in S_U$ as defined in Eq. (2). Let $M_v \in \mathbb{R}^{(101\times72)\times1}$ denote the vectorized form of $M$. A Euclidean coordinate matrix $X \in \mathbb{R}^{(101\times72)\times2}$ containing the slope-phase pair is defined below, over the same domain as the stacked sample matrix $M_v$.

$$x_i = \begin{bmatrix} s_i/35 \\ p_i \end{bmatrix} \in \mathbb{R}^2$$

$$X = \begin{bmatrix} x_1^T \\ x_2^T \\ \vdots \\ x_{7272}^T \end{bmatrix} = \begin{bmatrix} +1 & 0 \\ +1 & 0.01 \\ \vdots & \vdots \\ +1 & 1 \\ +34/35 & 0 \\ \vdots & \vdots \\ -1 & 1 \end{bmatrix} \in \mathbb{R}^{7272\times2} \tag{A2}$$

Using the Gaussian kernel matrix defined in Section II-C, the kernel matrix $\psi \in \mathbb{R}^{7272\times7272}$ was constructed from all sampled coordinate pairs. The RBF coefficient vector $A = [a_1 \; a_2 \; \ldots \; a_{7272}]^T \in \mathbb{R}^{7272\times1}$ were identified by solving

$$M_v = \psi A \tag{A3}$$

Because $\psi$ is singular, the RBF coefficients were obtained using the regularized solution described in Eq. (6).

To construct a mode-unified biomechanics surface, the RBF representation was subsequently evaluated on a unified slope-phase domain defined by Eqs. (2) and (3), without duplication. Similar to Eq. (A2), a Euclidean coordinate matrix $X_U \in \mathbb{R}^{7171\times2}$ corresponding to the unified and non-duplicated slope-phase domain consisting of $S_U$ and $P$ can be defined.

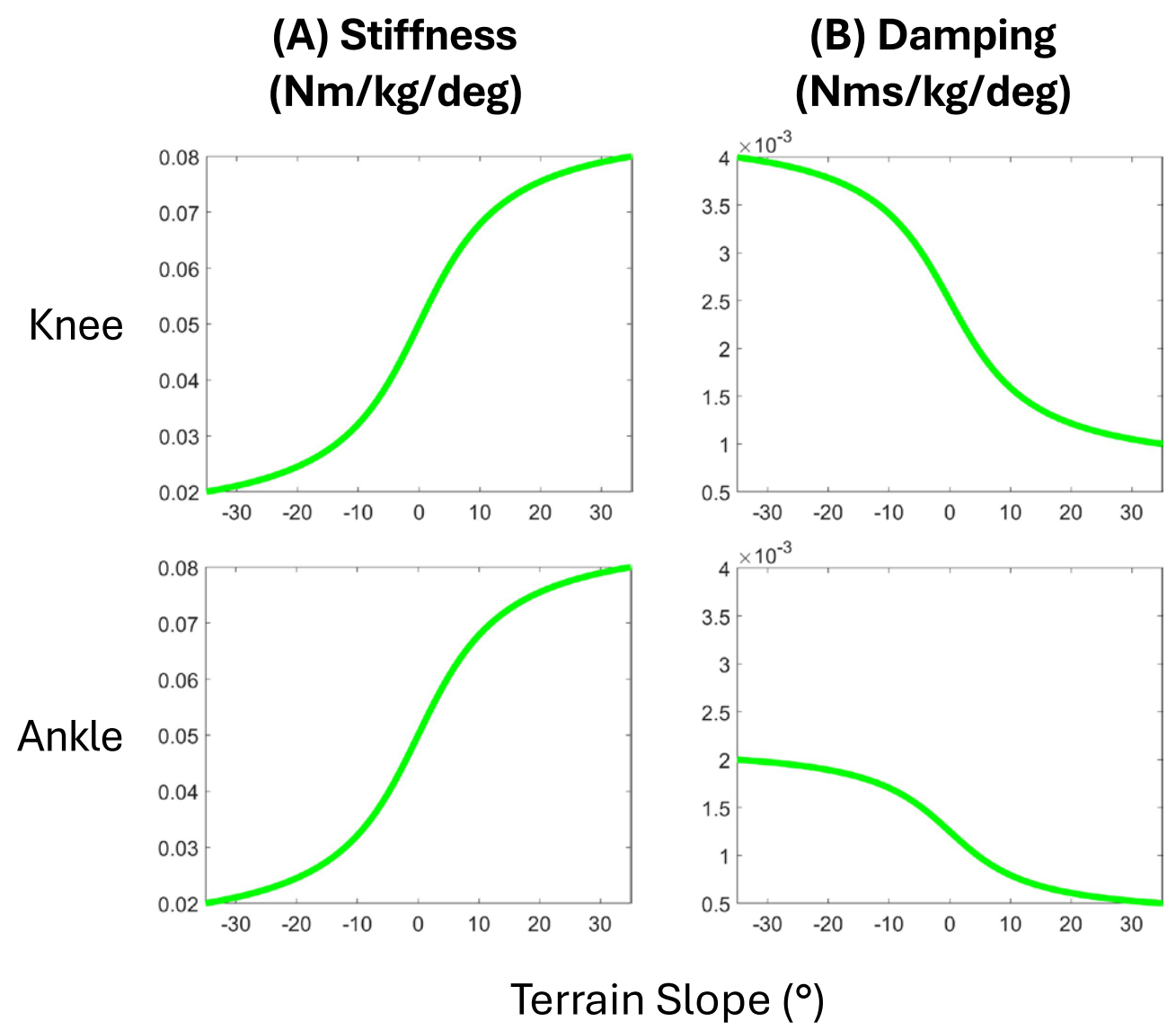

**Fig. A1.** Slope-dependent impedance bounds used for constrained least-squares optimization are illustrated. The arctangent upper envelopes for **(A)** stiffness and **(B)** damping for knee and ankle are shown across the unified terrain slope domain.

$$x_{U_i} = \begin{bmatrix} s_{U_i}/35 \\ p_i \end{bmatrix} \in \mathbb{R}^2$$

$$X_U = \begin{bmatrix} x_{U_1}^T \\ \vdots \\ x_{U_{7171}}^T \end{bmatrix} \tag{A4}$$

Then an interpolation kernel matrix $\psi_U$ was computed between the unified and the original sample coordinates $X_U$ and $X$ as

$$\psi_{U_{i,j}} = \exp\left(-\left|x_{U_i} - x_j\right|_2^2\right) \tag{A5}$$

where $x_{U_i}$ and $x_j$ denote the i-th and j-th coordinate of $X_U$ and $X$, respectively. Since the unified slope-phase domain contains 7171 query points and the original domain contains 7272 samples, each row of $\psi_U$ corresponds to a query coordinate in the unified slope-phase domain, while each column corresponds to an original sample coordinate used in Eq. (A3). Therefore, $\psi_U \in \mathbb{R}^{7171\times7272}$. The samples in the slope-phase domain were then obtained by evaluating the RBF representation using the coefficient vector $A \in \mathbb{R}^{7272\times1}$,

$$M_U = \psi_U A \tag{A6}$$

Finally, a unified model $M_U \in \mathbb{R}^{7171\times1}$ was reshaped into the slope-phase representation $f_U(s, p) \in \mathbb{R}^{71\times101}$, yielding a continuous and mode-unified biomechanics surface over the full slope domain, with no duplicated coordinate pair. The resulting continuous knee and ankle biomechanics representations were subsequently utilized for the impedance parameterization described in Section III.

The corresponding MATLAB package is publicly available in this GitHub Repository.

### B. Constrained Optimization of Impedance Parameters

The stance-phase stiffness and damping parameters were identified through constrained least-squares optimization, as introduced in Section III-B. The optimization was conducted independently for the knee and ankle, and separately for early-stance (ES) and late-stance (LS). This section describes the regression-matrix construction and constraints, and the selection of joint-specific ES-to-LS boundaries.

For each terrain slope, stiffness and damping were represented by fourth-order polynomials of normalized stance phase as defined in Eq. (10). Polynomial basis $\phi(p)$ and coefficient vector $\mathbf{x}^{\boldsymbol{q}}$ can be defined as

$$\phi(p) = [1 \quad p \quad p^2 \quad p^3 \quad p^4]^T,$$
$$\mathbf{x}^{\boldsymbol{q}} = \left[k_{s\,o}^{\boldsymbol{q}} \dots k_{s\,4}^{\boldsymbol{q}}\, b_{s\,o}^{\boldsymbol{q}} \dots b_{s\,4}^{\boldsymbol{q}}\right]^{\mathrm{T}} \in \mathbb{R}^{10\times 1}, \tag{A7}$$
$$\boldsymbol{q} \in \{ES, LS\}$$

The $j$-th row of the regression matrix $\boldsymbol{C}_{j,:}^{\boldsymbol{q}} \in \mathbb{R}^{1\times 10}$ was constructed as

$$\boldsymbol{C}_{j,:}^{\boldsymbol{q}} = \left[\Delta\theta_j \phi^T(p_j) \quad \dot{\theta}(s,p_j)\phi^T(p_j)\right]$$
$$= \left[\Delta\theta_j p_j^0 \dots \Delta\theta_j p_j^4 \quad \dot{\theta}(s,p_j)p_j^0 \dots \dot{\theta}(s,p_j)p_j^4\right], \tag{A8}$$
$$j \in \{0, 1, \dots, 100\}$$

where $\Delta\theta_j = \theta(s,p_j) - \theta_{eq}(s,p_j)$. The joint angle, angular velocity, and target moment were obtained from the RBF-unified biomechanics described in Section II. The resulting regression matrix was used in the constrained least-squares formulation in (11) to identify the polynomial coefficients independently over ES and LS.

To accommodate the varying mechanical demands across terrain slopes, the upper bounds of stiffness and damping were represented using continuous arctangent functions. The stiffness upper bound was designed to increase with terrain slope to accommodate greater joint support demands during ascent, whereas the damping upper bound decreased with slope to allow greater energy dissipation during descent. The arctangent models were fitted using reference values at ± 35 ° and 0 °, as illustrated in Fig. A1. Constant lower bounds were prescribed for both parameters and used as their initial values at heel contact. The numerical bounds are listed in Table A1, and the complete arctangent fitting procedure and coefficients are provided in the accompanying MATLAB repository.

Equality constraints were set to anchor the impedance trajectories at selected stance-phase reference points,

$$\boldsymbol{A}_{\mathrm{eq}}^{\boldsymbol{q}}\mathbf{x}^{\boldsymbol{q}} = \boldsymbol{\beta}_{\mathrm{eq}}^{\boldsymbol{q}}, \quad \boldsymbol{q} \in \{ES, LS\} \tag{A9}$$

ES begins with the minimum impedance values at heel-contact. Hence, impedances at the heel-contact were constrained by

$$\boldsymbol{A}_{\mathrm{eq}}^{\mathrm{ES}} = \begin{bmatrix} \phi^T(0) & \mathbf{0} \\ \mathbf{0} & \phi^T(0) \end{bmatrix}, \qquad \boldsymbol{\beta}_{\mathrm{eq}}^{\mathrm{ES}} = \begin{bmatrix} K_{\min} \\ B_{\min} \end{bmatrix} \tag{A10}$$

**Table A1.** Joint-specific phase references, impedance bounds, and values for shape constraint coefficients.

| Notation | Parameter | Knee | Ankle |
|---|---|---|---|
| $\boldsymbol{n_{ES}}$ | Number of ES samples | 65 | 55 |
| $\boldsymbol{n_{LS}}$ | Number of LS samples | 35 | 45 |
| $\boldsymbol{p_{ES}}$ | Terminal ES phase | 0.65 | 0.55 |
| $\boldsymbol{p_{LS}}$ | Initial LS phase | 0.66 | 0.56 |
| $\boldsymbol{p_K}$ | LS stiffness-reference phase | 0.75 | 0.75 |
| $\boldsymbol{K_{max}}(+35\,°)$ | Stiffness upper envelope at **+35** ° | 0.080 | 0.080 |
| $\boldsymbol{K_{max}}(-35\,°)$ | Stiffness upper envelope at **−35** ° | 0.020 | 0.020 |
| $\boldsymbol{K_{min}}$ | Stiffness lower bound | 0.008 | 0.016 |
| $\boldsymbol{B_{max}}(+35\,°)$ | Damping upper envelope at **+35** ° | 0.001 | 0.0005 |
| $\boldsymbol{B_{max}}(-35\,°)$ | Damping upper envelope at **−35** ° | 0.004 | 0.002 |
| $\boldsymbol{B_{min}}$ | Damping lower bound | 0.0004 | 0.0004 |
| $\boldsymbol{\varepsilon_{ES_1}}$ | Signed right-hand-side constraint of $\dot{K}$ during $ES_1$ | −0.010 | +0.010 |
| $\boldsymbol{\varepsilon_{ES_2}}$ | Signed right-hand-side constraint of $\dot{B}$ during $ES_1$ | −0.001 | −0.001 |
| $\boldsymbol{\varepsilon_{ES_3}}$ | Signed right-hand-side constraint of $\dot{K}$ during $ES_2$ | +0.010 | +0.010 |
| $\boldsymbol{\varepsilon_{ES_4}}$ | Signed right-hand-side constraint of $\dot{K}$ during $ES_2$ | +0.001 | +0.001 |
| $\boldsymbol{\varepsilon_{ES_5}}$ | Signed right-hand-side constraint of $\dot{B}$ during $ES_2$ | +0.010 | +0.005 |
| $\boldsymbol{\varepsilon_{ES_6}}$ | Signed right-hand-side constraint of $\dot{B}$ during $ES_2$ | +0.001 | +0.001 |
| $\boldsymbol{\varepsilon_{LS_1}}$ | Signed right-hand-side constraint of $\dot{K}$ during $LS_1$ | +0.005 | +0.005 |
| $\boldsymbol{\varepsilon_{LS_2}}$ | Signed right-hand-side constraint of $\dot{K}$ during $LS_2$ | +0.0001 | +0.00001 |
| $\boldsymbol{\varepsilon_{LS_3}}$ | Signed right-hand-side constraint of $\dot{B}$ during $LS_2$ | +0.00001 | 0 |

The LS optimization was initialized using the terminal ES solution and constrained to preserve the support-level impedance until the onset of unloading, and return to the minimum values at toe-off. Denoting $p_{ES}$, $p_{LS}$, and $p_K$ as the terminal ES, initial LS, and reference point during LS yields

$$\boldsymbol{A}_{\mathrm{eq}}^{\mathrm{LS}} = \begin{bmatrix} \phi^T(p_{LS}) & \mathbf{0} \\ \phi^T(1) & \mathbf{0} \\ \mathbf{0} & \phi^T(p_{LS}) \\ \mathbf{0} & \phi^T(1) \\ \phi^T(p_K) & 0 \end{bmatrix}, \quad \boldsymbol{\beta}_{\mathrm{eq}}^{\mathrm{LS}} = \begin{bmatrix} K_{\mathrm{ES}}(s,p_{ES}) \\ K_{\min} \\ B_{\mathrm{ES}}(s,p_{ES}) \\ B_{\min} \\ K_{\mathrm{ES}}(s,p_{ES}) \end{bmatrix} \tag{A11}$$

An additional equality constraint was imposed on stiffness near

75% of stance by $p_K$, to preserve the support-level impedance before unloading. In contrast, damping was allowed to vary within the prescribed bounds and derivative constraints, without an additional equality constraint at this reference point. This distinction reflects the different mechanical roles of stiffness and damping in regulating position-dependent joint support and velocity-dependent resistance, respectively. For the knee, independent damping modulation also accommodates controlled yielding and preparation for swing initiation, consistent with previous prosthesis control strategies (Fey *et al.*, 2014). Similarly, previous studies of biological ankle impedance have reported distinct stance-phase modulation of stiffness and damping (Rouse *et al.*, 2014; Lee, Rouse and Krebs, 2016; Shorter and Rouse, 2018). The reference points were defined separately for each joint, as detailed in Table A1.

The remaining portions of the optimization were formulated as inequality constraints with bound and shape components,

$$\boldsymbol{A}^{\boldsymbol{q}}_{\text{ineq}}\mathbf{x}^{\mathbf{q}} \le \boldsymbol{\beta}^{\boldsymbol{q}}_{\text{ineq}}$$

$$\boldsymbol{A}^{\boldsymbol{q}}_{\text{ineq}} = \begin{bmatrix} \boldsymbol{A}^{\boldsymbol{q}}_{\text{bound}} \\ \boldsymbol{A}^{\boldsymbol{q}}_{\text{shape}} \end{bmatrix}, \qquad \boldsymbol{\beta}^{\boldsymbol{q}}_{\text{ineq}} = \begin{bmatrix} \boldsymbol{\beta}^{\boldsymbol{q}}_{\text{bound}} \\ \boldsymbol{\beta}^{\boldsymbol{q}}_{\text{shape}} \end{bmatrix}, \tag{A12}$$

$$\boldsymbol{q} \in \{ES, LS\}$$

The bound constraints enforce the admissible stiffness and damping envelopes in Eq. (12) throughout the corresponding optimization interval. Defining $\mathcal{P}_{ES} = [\phi(0) \dots \phi(p_{ES})]^T$, $\mathcal{P}_{LS} = [\phi(p_{LS}) \dots \phi(1)]^T$, and $\boldsymbol{l}_j = [1 \dots 1]^T \in \mathbb{R}^{j\times 1}$, the corresponding bound matrices are

$$\boldsymbol{A}^{\text{ES}}_{\text{bound}} = \begin{bmatrix} \mathcal{P}_{\text{ES}} & \mathbf{0} \\ -\mathcal{P}_{\text{ES}} & \mathbf{0} \\ \mathbf{0} & \mathcal{P}_{\text{ES}} \\ \mathbf{0} & -\mathcal{P}_{\text{ES}} \end{bmatrix}, \qquad \boldsymbol{\beta}^{\text{ES}}_{\text{bound}} = \begin{bmatrix} K_{\max}(s)\boldsymbol{l}_{n_{ES}} \\ -K_{\min}\boldsymbol{l}_{n_{ES}} \\ B_{\max}(s)\boldsymbol{l}_{n_{ES}} \\ -B_{\min}\boldsymbol{l}_{n_{ES}} \end{bmatrix}$$

$$\boldsymbol{A}^{\text{LS}}_{\text{bound}} = \begin{bmatrix} \mathcal{P}_{\text{LS}} & \mathbf{0} \\ -\mathcal{P}_{\text{LS}} & \mathbf{0} \\ \mathbf{0} & \mathcal{P}_{\text{LS}} \\ \mathbf{0} & -\mathcal{P}_{\text{LS}} \end{bmatrix}, \qquad \boldsymbol{\beta}^{\text{LS}}_{\text{bound}} = \begin{bmatrix} K_{ES}(s,p_{ES})\boldsymbol{l}_{n_{LS}} \\ -K_{\min}\boldsymbol{l}_{n_{LS}} \\ B_{ES}(s,p_{ES})\boldsymbol{l}_{n_{LS}} \\ -B_{\min}\boldsymbol{l}_{n_{LS}} \end{bmatrix} \tag{A13}$$

These bound constraints prevent the LS solution from exceeding the support-level impedance established during ES and preserves the intended plateau-to-unloading structure.

The shape constraints are then constructed from the derivative polynomial basis $\dot{\phi}(p) = [0 \quad 1 \quad 2p \quad 3p^2 \quad 4p^3]^T$ to regulate the impedance progression within the envelopes. Accordingly, the derivatives of $K(s,p)$ and $B(s,p)$ with respect to normalized stance phase are represented as

$$\frac{\partial K(s,p)}{\partial p} = \sum_{i=1}^{4} i k_{s_i} p^{i-1}, \tag{A14}$$

$$\frac{\partial B(s,p)}{\partial p} = \sum_{i=1}^{4} i b_{s_i} p^{i-1}$$

Defining derivative terms over the ES and LS intervals as $\dot{\mathcal{P}}_{ES_1} = [\dot{\phi}(0) \dots \dot{\phi}(0.25)]^T$, $\dot{\mathcal{P}}_{ES_2} = [\dot{\phi}(0.26) \dots \dot{\phi}(p_{ES})]^T$, $\dot{\mathcal{P}}_{LS_1} = [\dot{\phi}(p_{LS}) \dots \dot{\phi}(p_K)]^T$, and $\dot{\mathcal{P}}_{LS_2} = [\dot{\phi}(p_K) \dots \dot{\phi}(1)]^T$ yields

$$\boldsymbol{A}^{\text{ES}}_{\text{shape}} = \begin{bmatrix} -\dot{\mathcal{P}}_{\text{ES}_1} & \mathbf{0} \\ \mathbf{0} & -\dot{\mathcal{P}}_{\text{ES}_1} \\ \dot{\mathcal{P}}_{\text{ES}_2} & \mathbf{0} \\ -\dot{\mathcal{P}}_{\text{ES}_2} & \mathbf{0} \\ \mathbf{0} & \dot{\mathcal{P}}_{\text{ES}_2} \\ \mathbf{0} & -\dot{\mathcal{P}}_{\text{ES}_2} \end{bmatrix}, \qquad \boldsymbol{\beta}^{\text{ES}}_{\text{shape}} = \begin{bmatrix} \varepsilon_{ES_1}\boldsymbol{l}_{n_1} \\ \varepsilon_{ES_2}\boldsymbol{l}_{n_1} \\ \varepsilon_{ES_3}\boldsymbol{l}_{n_2} \\ \varepsilon_{ES_4}\boldsymbol{l}_{n_2} \\ \varepsilon_{ES_5}\boldsymbol{l}_{n_2} \\ \varepsilon_{ES_6}\boldsymbol{l}_{n_2} \end{bmatrix} \tag{A15}$$

$$\boldsymbol{A}^{\text{LS}}_{\text{shape}} = \begin{bmatrix} -\dot{\mathcal{P}}_{\text{LS}_1} & \mathbf{0} \\ \dot{\mathcal{P}}_{\text{LS}_2} & \mathbf{0} \\ \mathbf{0} & \dot{\mathcal{P}}_{\text{LS}_2} \end{bmatrix}, \qquad \boldsymbol{\beta}^{\text{LS}}_{\text{shape}} = \begin{bmatrix} \varepsilon_{LS_1}\boldsymbol{l}_{n_3} \\ \varepsilon_{LS_2}\boldsymbol{l}_{n_4} \\ \varepsilon_{LS_3}\boldsymbol{l}_{n_4} \end{bmatrix}$$

Here, $n_1 = 25$, $n_2 = n_{ES} - 25$, where $n_{ES}$ is the number of sample points of ES. $ES_1$ ends at sample 25 and $ES_2$ begins at sample 26. In addition, $n_3 = 76 - n_{ES}$ and $n_4 = 25$. Accordingly, the first LS shape constraint limits the minimum stiffness derivative throughout $LS_1$ ($p_{LS}$ to $p_K$), while the remaining constraints restrict increases in stiffness and damping during unloading ($p_K$ to TO). The detailed values for coefficients used in matrix representation are listed in Table A1. Stiffness and damping are expressed in Nm/kg/deg and Nms/kg/deg, respectively. Phase values are normalized to stance; $p_K \approx 0.75$. The $\varepsilon$ coefficients are signed right-hand-side values of the inequalities in Eq. (A15) and the subsequent LS shape constraints. Note that these are offline optimization settings, not the real-time FSM transition timing.

The complete MATLAB package, including the constrained impedance parameterization and evaluation scripts are available in this GitHub Repository. Above optimization settings are defined in src/config/get_joint_config.m, allowing readers to evaluate alternative combinations of phase boundaries, impedance bounds, and derivative limits.

### *C. Pseudo-code for Controller Implementation*

The continuous impedance parameters derived in Section III were implemented within a unified, load-reactive FSM controller. As summarized in Algorithm A1, the controller first receives terrain slope information and selects the corresponding knee and ankle stiffness, damping, and equilibrium-angle references from the unified parameter representations illustrated in Fig. 7. These selected parameters are modulated according to prosthetic-side vertical loading and the current FSM state.

During early stance (ES), stiffness and damping progressively increase with prosthetic loading toward the selected support values, while the equilibrium angles follow the prescribed joint-specific strategies described in Section III-A. Once the loading reaches the upper threshold, the impedance

**Algorithm A1.** Pseudocode for the mode-unified FSM implementation. Terrain slope information selects knee and ankle impedance references from the unified parameter representations.

***High-Level***
1: FSM receives slope information
→ selects slope-dependent impedance parameter trajectories

***Early Stance***
1: if Load-cell force Z ≤ 50% body weight
→ $K$, $B$, and $\theta_{eq}$ progress along increasing load
*** Figure A2-a**
2: if Load-cell force Z ≥ 50% body weight
→ $K$, $B$, and $\theta_{eq}$ are anchored
3: if Load-cell force Z ≥ 50% body weight & intact ankle is forward **(+)**
→ enter ***Late Stance***
*** Figure A2-b**

***Late Stance***
4: if Load-cell force Z ≥ 50% body weight
→ $K$, $B$, and $\theta_{eq}$ are anchored
5: if Load-cell force Z ≤ 50% body weight
→ $K$, $B$, and $\theta_{eq}$ progress along decreasing load
*** Figure A2-c**
6: if Load-cell force Z ≤ 10% body weight
→ enter ***Swing Flexion***

***Swing Flexion***
7: Go to maximum flexion (knee flexion, ankle plantarflexion)
8: if Knee angular velocity ≤ 1 ° / sec
→ enter ***Swing Extension***

***Swing Extension***
9: Go to heel contact angle
10: if Load-cell force Z ≥ 10% body weight
→ enter ***Early Stance***

parameters are maintained throughout ES. The ES-to-LS transition is then triggered by the bilateral limb-progression criterion described in Section IV-B, subject to sufficient prosthetic loading. During late stance (LS), the controller maintains joint support before progressively modulating the impedance parameters with decreasing load to facilitate unloading, ankle push-off, and knee swing initiation. The swing flexion and swing extension states employ swing impedance parameters and slope-dependent equilibrium angles to execute ground clearance and preparation for heel contact.

Figure A2 illustrates a representative gait cycle, showing the relationship between bilateral limb progression, prosthetic loading, FSM transitions, and the resulting knee and ankle impedance parameters. During initial weight acceptance (Fig. A2a), stiffness and damping increase with loading. The intact ankle subsequently advances beyond the prosthetic ankle, triggering the ES-to-LS transition under sufficient load (Fig. A2b). As the prosthetic limb unloads (Fig. A2c), the controller progresses toward the terminal stance and swing behaviors. These transitions are implemented within the same FSM structure across locomotion conditions.

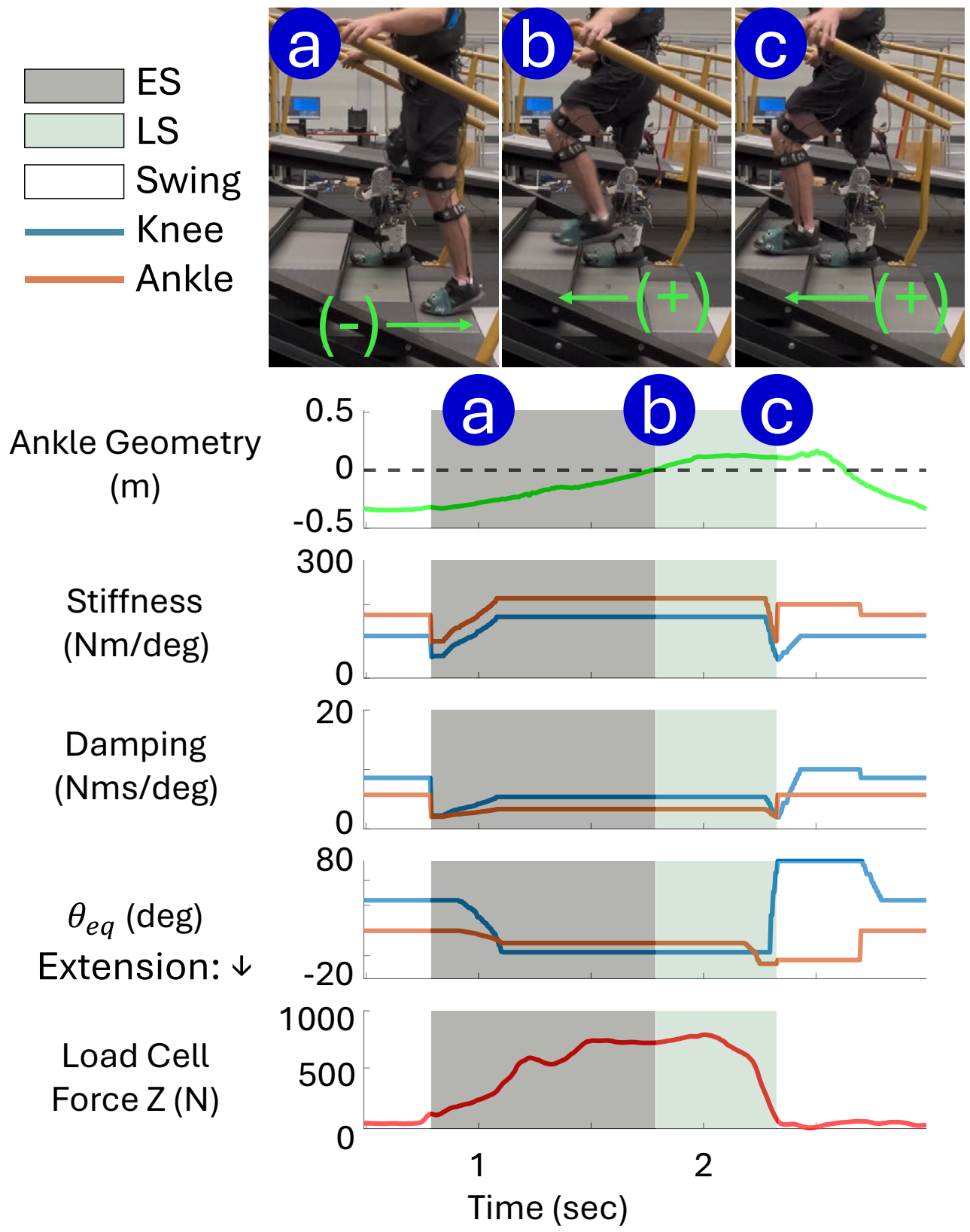

**Fig. A2.** Representative implementation of proposed controller. Bilateral ankle progression, prosthetic stiffness, damping, equilibrium angle, and vertical load are illustrated throughout a gait cycle. Shaded regions identify early stance (ES), late stance (LS), and swing. Representative instants (a, b, and c) illustrate increasing prosthetic loading during weight acceptance, the bilateral limb crossing event associated with the ES-to-LS transition and unloading. Knee and ankle impedance parameters evolved according to the current FSM state and measured prosthetic loading (load-cell force Z).

### *D. Comparison with a Previous Mode-specific Controller*

The comparison in Fig. 11 examined biomechanical outcomes from the previously reported mode-specific controller (n = 9) (Zhou *et al.*, 2025) and the proposed unified controller (n = 10) across four common terrain conditions (± 5 ° ramps and ± 24 ° stairs). The same biomechanical landmarks evaluated in Figs. 10A and 10B, including heel-contact, swing-clearance, and toe-off angles, together with peak stance torque, were extracted from both datasets. Subject-level errors relative to the AB references were calculated, and between-study differences in these errors were assessed using Welch's two-sample t-tests to account for unequal size and potentially unequal variances (*$p < 0.05$, **$p < 0.01$, ***$p < 0.001$, ****$p < 0.0001$).

Figure 11 presents the corresponding experimental outcomes alongside the AB reference, with annotations highlighting the differences in mean absolute error (MAE) between studies. Among the 24 evaluated comparisons, six exhibited significantly lower MAEs in the present study. Knee swing clearance MAE during ramp ascent decreased from 27.1 ° to 6.6 ° ($p < 0.0001$). Ankle heel-contact MAE decreased from 14.2 ° to 2.2 ° during stair ascent ($p < 0.01$) and from 19.3 ° to 5.8 ° during stair descent ($p < 0.0001$). Ankle swing-clearance MAE during stair ascent decreased from 12.6 ° to 2.4 ° ($p < 0.01$), while ankle toe-off MAE during ramp descent decreased from 16.3 ° to 2.5 ° ($p < 0.0001$). Knee peak stance-torque MAE during stair descent decreased from 0.52 to 0.28 Nm/kg ($p < 0.05$).

Several additional outcomes showed lower mean MAEs with the proposed controller, although the between-group differences did not reach statistical significance. For example, knee toe-off angle MAE decreased from 8.81 ° to 4.38 ° during ramp descent and from 11.60 ° to 6.28 ° during stair descent, while ankle heel-contact angle MAE during ramp ascent decreased from 5.08 ° to 2.55 °. Knee peak stance-torque MAE also decreased from 0.33 to 0.18 Nm/kg during ramp descent. Lower errors were additionally observed for knee swing clearance angle and knee peak stance torque during stair ascent. In contrast, ankle peak stance torque did not show a consistent improvement across terrain conditions, consistent with the comparatively weaker reproduction of ankle kinetics identified for the proposed controller in Fig. 10B.

To further examine whether differences in prosthetic joint behavior were accompanied by changes in limb loading, weight-normalized vGRF profiles were visually compared during the matched overground stair conditions (Fig. 11B). These comparisons were restricted to stairs, for which both controller groups were tested overground, to characterize differences in limb loading. During stair descent, prosthetic- and intact-side loading remained comparable between controllers. During stair ascent, however, the proposed controller produced a higher prosthetic-side vGRF than the previously reported controller, whereas intact-side loading remained comparable.